\documentclass{article} 
\usepackage{iclr2027_conference,times}

\usepackage{amsmath,amsfonts,bm}

\def\eqref#1{equation~\ref{#1}}

\def\1{\bm{1}}

\def\ve{{\bm{e}}}

\def\vg{{\bm{g}}}
\def\vh{{\bm{h}}}

\def\vw{{\bm{w}}}

\def\vz{{\bm{z}}}

\def\mA{{\bm{A}}}
\def\mB{{\bm{B}}}

\def\mD{{\bm{D}}}
\def\mE{{\bm{E}}}

\def\mH{{\bm{H}}}
\def\mI{{\bm{I}}}

\def\mL{{\bm{L}}}
\def\mM{{\bm{M}}}

\def\mP{{\bm{P}}}

\def\mW{{\bm{W}}}

\def\mZ{{\bm{Z}}}

\DeclareMathAlphabet{\mathsfit}{\encodingdefault}{\sfdefault}{m}{sl}
\SetMathAlphabet{\mathsfit}{bold}{\encodingdefault}{\sfdefault}{bx}{n}

\newcommand{\R}{\mathbb{R}}

\usepackage[utf8]{inputenc}
\usepackage[T1]{fontenc}
\usepackage{hyperref}
\usepackage{url}
\usepackage{booktabs}
\usepackage{amsmath,amssymb}
\usepackage{graphicx}
\usepackage{fix-cm}
\usepackage{pgfplots}
\pgfplotsset{compat=1.18}
\usepackage{xcolor}
\usepackage{multirow}
\usepackage{pifont}
\usepackage{listings}
\usepackage{enumitem}
\usepackage{microtype}

\lstdefinestyle{python}{
  basicstyle=\ttfamily\scriptsize,
  keywordstyle=\color{blue!60!black},
  commentstyle=\color{black!55}\itshape,
  columns=fullflexible,
  keepspaces=true,
  showstringspaces=false,
  language=Python,
  aboveskip=2pt,
  belowskip=2pt,
}

\newcommand{\Twin}{Twin embeddings}
\newcommand{\method}{on\'{o}na}
\newcommand{\gate}{\method{}}  

\title{\method{}: Twin Embeddings\\with Tied Unembeddings}

\author{\makebox[\dimexpr0.32\textwidth-2\tabcolsep\relax][l]{Cheng Luo} \\
{\fontsize{7.8}{9.5}\selectfont Independent Researcher} \\
{\fontsize{7}{9}\selectfont\texttt{wdlctc@gmail.com}}
\And
\makebox[\dimexpr0.32\textwidth-2\tabcolsep\relax][l]{Xinnan Dai} \\
{\fontsize{7.8}{9.5}\selectfont Michigan State University} \\
{\fontsize{7}{9}\selectfont\texttt{daixinna@msu.edu}}
\And
\makebox[\dimexpr0.32\textwidth-2\tabcolsep\relax][l]{Zhongzhu Zhou} \\
{\fontsize{7.8}{9.5}\selectfont Together AI} \\
{\fontsize{7}{9}\selectfont\texttt{zhongzhu.zhou@sydney.edu.au}}
\AND
\makebox[\dimexpr0.32\textwidth-2\tabcolsep\relax][l]{Zefan Cai} \\
{\fontsize{7.8}{9.5}\selectfont University of Wisconsin--Madison} \\
{\fontsize{7}{9}\selectfont\texttt{zefncai@gmail.com}}
\And
\makebox[\dimexpr0.32\textwidth-2\tabcolsep\relax][l]{Ruijie Zhu} \\
{\fontsize{7.8}{9.5}\selectfont University of California, Santa Cruz} \\
{\fontsize{7}{9}\selectfont\texttt{ridger@ucsc.edu}}
\And
\makebox[\dimexpr0.32\textwidth-2\tabcolsep\relax][l]{Tianyu Zhang} \\
{\fontsize{7.8}{9.5}\selectfont Mila} \\
{\fontsize{7}{9}\selectfont\texttt{tianyu.zhang@mila.quebec}}
\AND
\makebox[\dimexpr\textwidth-2\tabcolsep\relax][c]{Junjie Hu}\\
\makebox[\dimexpr\textwidth-2\tabcolsep\relax][c]{{\fontsize{7.8}{9.5}\selectfont University of Wisconsin--Madison}}\\
\makebox[\dimexpr\textwidth-2\tabcolsep\relax][c]{{\fontsize{7}{9}\selectfont\texttt{junjie.hu@wisc.edu}}}}

\iclrfinalcopy
\hypersetup{
  pdftitle={on\'{o}na: Twin Embeddings with Tied Unembeddings},
  pdfauthor={Cheng Luo, Xinnan Dai, Zhongzhu Zhou, Zefan Cai, Ruijie Zhu, Tianyu Zhang, Junjie Hu}
}
\begin{document}
\maketitle
\fancyhead{}
\renewcommand{\headrulewidth}{0pt}

\begin{abstract}
Weight tying reduces the parameter cost of language models, but the usual tied output head is limited to a single softmax. We introduce \method{}, a two-table parameterization that increases output expressiveness while preserving input--output weight sharing. The Transformer takes the average of two embedding tables as input, and the output head combines the softmax distributions defined by those same tables. A gate conditioned only on the current input token determines the mixture weight. Its values can be precomputed for inference, and the Transformer body remains unchanged. We evaluate \method{} across Dense, Gated DeltaNet hybrid, mixture-of-experts, and Attention Residual architectures at 142M, 330M, and 1B scales. Under matched token budgets, \method{} lowers final validation loss and external perplexity in 11 of 12 comparisons and raises the average accuracy over eight downstream tasks in 10 of 12. The one exception on loss is the 1B MoE model. The cost is one extra embedding table and one extra vocabulary projection.

\end{abstract}

\section{Introduction}
\label{sec:intro}

Weight tying uses the same table for the input embedding and output layer of a language model~\citep{press2017using,inan2017tying,liu2024mobilellm,yang2025qwen3}. The table $\mE\in\R^{V\times d}$ maps each input token to a vector and, in its role as the \emph{unembedding}, scores every vocabulary entry as a possible next token. Sharing these roles saves a $V\times d$ matrix, a substantial fraction of a small model's parameters. Tying also improved perplexity in the original experiments~\citep{press2017using}, and the fully untied 400M control in \citet{batley2026leviathan} underperformed its tied baseline. Both roles contribute to learning, but early in training the tied matrix is shaped mostly by output gradients~\citep{lopardo2026weight}.

The usual tied head scores the vocabulary with a single table followed by a softmax. Its matrix of log-probabilities across contexts has rank at most $d+1$, limiting its ability to represent high-rank next-token distributions. This restriction is known as the \emph{softmax bottleneck}~\citep{yang2018breaking}. It matters when the next token has several plausible modes~\citep{chang2022softmax} and has been linked to the saturation of small models~\citep{godey2024why}. A mixture of softmaxes (MoS) addresses the restriction by combining predictions from $K$ transformed contexts under one table, at a cost of $K$ vocabulary projections per position~\citep{yang2018breaking}. 

Two obvious fixes are to untie the output table or to add a second softmax component under the tied table. Neither works well in our controls. An untied head adds $V\times d$ parameters but gives up input--output sharing, and finishes 0.009 nats above the tied baseline at 142M. A $K{=}2$ MoS keeps sharing, but every component sends its gradient into the same input table, and it finishes 0.024 nats above the baseline. We want instead a head that gains capacity while every table used to score the vocabulary stays tied to the input.

\textbf{\Twin{} with tied unembeddings} implement this design with two tables of the same shape, $\mA$ and $\mB$ (Figure~\ref{fig:method}). The body reads their average $\mM=\frac12(\mA+\mB)$, which serves as a single input table. The head scores the vocabulary with both tables and mixes the resulting distributions using a weight $g_t$ that depends only on the input token $x_t$. We call this model \method{}. Whereas MoS combines $K$ contexts under one table, \method{} combines two tables under one context. It adds a second table and $d$ gate weights, and each tied table receives its own output-projection gradient.

Conditioning the gate on token identity allows its values to be computed once for the whole vocabulary, following the use of token identity in hash-based expert routing~\citep{roller2021hash}. At serving time, the input tables fold into $\mM$ and the gate becomes a lookup in a vector of $V$ values. The output still requires two vocabulary projections. 

We train Qwen3-style decoders on ClimbMix~\citep{diao2025climb} at 142M, 330M and 1B. At 142M, \gate{} lowers final validation loss by 0.0285 nats for Dense, 0.0134 for the Gated DeltaNet hybrid, 0.0116 for MoE and 0.0226 for Attention Residuals. At 330M, the corresponding improvements are 0.0065, 0.0037, 0.0144 and 0.0119 nats. Both external perplexity benchmarks improve in all eight pairs, and all sixteen one-table ablations have higher loss than their mixture (Table~\ref{tab:main}). The gains extend to the 1B Dense, GDN hybrid and AttnRes models. The 1B MoE model, however, has worse validation loss and perplexity despite a higher downstream average (Table~\ref{tab:downstream}).

\begin{figure}[t]
\centering
\setlength{\abovecaptionskip}{2pt}
\begin{minipage}[c]{0.32\textwidth}\centering
\begingroup
\usetikzlibrary{arrows.meta}
\definecolor{methodLine}{HTML}{4B515A}
\definecolor{methodGray}{HTML}{F1F2F4}
\definecolor{methodEdge}{HTML}{777B82}
\definecolor{methodGroup}{HTML}{B5BDC8}
\definecolor{methodLabel}{HTML}{56616E}
\definecolor{methodBody}{HTML}{E4D9EC}
\definecolor{methodBodyBack}{HTML}{F3EDF7}
\definecolor{methodPurple}{HTML}{9B81B1}
\definecolor{methodPurpleBack}{HTML}{C0ACCD}
\definecolor{methodPurpleText}{HTML}{49385B}
\definecolor{methodAvg}{HTML}{FFF1BE}
\definecolor{methodAvgEdge}{HTML}{C9A744}
\definecolor{methodAvgText}{HTML}{705C27}
\definecolor{methodGate}{HTML}{F3D0D9}
\definecolor{methodGateEdge}{HTML}{BA788C}
\definecolor{methodGateText}{HTML}{784252}
\definecolor{methodRouter}{HTML}{70747B}
\resizebox{\linewidth}{!}{%
\begin{tikzpicture}[x=1cm,y=-1cm,line cap=round,line join=round,
 font=\fontfamily{phv}\selectfont,text=methodLine,
 wire/.style={draw=methodLine,line width=0.5pt},
 flow/.style={wire,-{Latex[length=1.2mm,width=0.8mm]}},
 projection/.style={draw=methodEdge,fill=methodGray,line width=0.5pt},
 label/.style={font=\fontfamily{phv}\fontsize{7.8}{9}\selectfont,inner sep=0pt},
 parameter/.style={font=\fontsize{10}{11}\selectfont,inner sep=0pt}]
\path[use as bounding box] (0.1,0.3) rectangle (6.55,9.2);
\draw[methodGroup,line width=0.35pt,rounded corners=1.1mm] (1.28,2.25) rectangle (6.32,3.70);
\draw[methodGroup,line width=0.35pt,rounded corners=1.1mm] (1.28,6.70) rectangle (6.32,8.15);
\node[font=\fontfamily{phv}\bfseries\fontsize{7.4}{8.5}\selectfont,text=methodLabel] at (3.8,2.48) {Twin Unembedding};
\node[font=\fontfamily{phv}\bfseries\fontsize{7.4}{8.5}\selectfont,text=methodLabel] at (3.8,6.93) {Twin Embedding};
\draw[wire] (3.8,8.75)--(3.8,8.43);
\draw[flow] (3.8,8.43)--(2.55,8.43)--(2.55,7.95);
\draw[flow] (3.8,8.43)--(5.05,8.43)--(5.05,7.95);
\draw[wire] (2.55,7.15)--(2.55,6.50)--(5.05,6.50)--(5.05,7.15);
\draw[flow] (3.8,6.50)--(3.8,6.20);
\draw[flow] (3.8,5.60)--(3.8,4.77);
\draw[wire] (3.8,3.91)--(3.8,3.85);
\draw[flow] (3.8,3.85)--(2.55,3.85)--(2.55,3.50);
\draw[flow] (3.8,3.85)--(5.05,3.85)--(5.05,3.50);
\foreach \x/\arm in {2.55/A,5.05/B}{
  \path[projection] (\x-0.68,7.15)--(\x+0.68,7.15)--(\x+1,7.95)--(\x-1,7.95)--cycle;
  \node[label] at (\x,7.40) {Embedding};
  \node[parameter] at (\x,7.69) {$\boldsymbol{\arm}$};
  \path[projection] (\x-1,2.70)--(\x+1,2.70)--(\x+0.68,3.50)--(\x-0.68,3.50)--cycle;
  \node[label,font=\fontfamily{phv}\fontsize{7.4}{8.5}\selectfont] at (\x,2.95) {Unembedding};
  \node[parameter] at (\x,3.24) {$\boldsymbol{\arm}$};
}
\foreach \d in {0.14,0.07}{
 \path[draw=methodPurpleBack,fill=methodBodyBack,line width=0.5pt,rounded corners=0.85mm]
 (1.55+\d,4.05-\d) rectangle (6.05+\d,4.77-\d);
}
\path[draw=methodPurple,fill=methodBody,line width=0.5pt,rounded corners=0.85mm] (1.55,4.05) rectangle (6.05,4.77);
\node[font=\fontfamily{phv}\fontsize{9.2}{10.5}\selectfont,text=methodPurpleText] at (3.8,4.41) {Transformer};
\draw[draw=methodRouter,line width=0.5pt,dash pattern=on 1.6pt off 1.4pt,-{Latex[length=1.2mm,width=0.8mm]}]
 (3.5,5.90)--(0.62,5.90)--(0.62,1.10)--(3.5,1.10);
\node[label,anchor=west] at (0.90,5.28) {Router};
\node[circle,draw=methodAvgEdge,fill=methodAvg,line width=0.5pt,minimum size=0.60cm,inner sep=0pt,
 font=\fontfamily{phv}\fontsize{7.7}{9}\selectfont,text=methodAvgText] at (3.8,5.90) {avg};
\draw[wire] (2.55,2.70)--(2.55,1.96)--(5.05,1.96)--(5.05,2.70);
\draw[flow] (3.8,1.96)--(3.8,1.40);
\node[circle,draw=methodGateEdge,fill=methodGate,line width=0.5pt,minimum size=0.60cm,inner sep=0pt,
 font=\fontsize{10.5}{12}\selectfont,text=methodGateText] at (3.8,1.10) {$g_t$};
\draw[flow] (3.8,0.80)--(3.8,0.45);
\node[draw=methodGroup,fill=methodGray,line width=0.5pt,rounded corners=1.6mm,minimum width=0.72cm,minimum height=0.36cm,
 inner sep=0pt,font=\fontsize{9.3}{10}\selectfont] at (3.8,8.93) {$x_t$};
\end{tikzpicture}}
\endgroup\end{minipage}\hfill
\begin{minipage}[c]{0.67\textwidth}\centering
\begingroup
\definecolor{archDense}{HTML}{A8BBD9}
\definecolor{archGDN}{HTML}{ACD2B5}
\definecolor{archMoE}{HTML}{C8CEE4}
\definecolor{archAttn}{HTML}{EFBBD9}
\definecolor{curveStd}{HTML}{E07B39}
\definecolor{curveTwin}{HTML}{4B2C9B}
\resizebox{\linewidth}{!}{%
\begin{tikzpicture}[font=\sffamily]
\pgfplotsset{stackedbar/.style={scale only axis,width=2.42in,height=1.50in,
xmin=-0.6,xmax=3.6,ymin=-0.035,ymax=0.006,axis on top,
xtick={0,1,2,3},ytick={-0.03,-0.02,-0.01,0},scaled y ticks=false,
yticklabel style={/pgf/number format/fixed,/pgf/number format/precision=2,/pgf/number format/zerofill},
tick label style={font=\sffamily\fontsize{8.5}{9.5}\selectfont},
axis line style={gray,line width=0.5pt},tick style={black,line width=0.5pt},
clip=false,title style={font=\sffamily\bfseries\fontsize{10}{11}\selectfont,yshift=3pt}}}
\begin{axis}[stackedbar,at={(0,1.70in)},anchor=south west,xticklabels={,,,},xtick style={draw=none},title={(b) Final gain over Standard}]
\addplot[ybar,bar width=31.8pt,fill=archDense,draw=white,line width=0.4pt] coordinates {(0,-0.028479013592)};
\node[anchor=north,font=\sffamily\fontsize{8.5}{9.5}\selectfont,inner sep=0pt] at (axis cs:0,-0.029679013592) {$ -0.0285 $};
\addplot[ybar,bar width=31.8pt,fill=archGDN,draw=white,line width=0.4pt] coordinates {(1,-0.013359507546)};
\node[anchor=north,font=\sffamily\fontsize{8.5}{9.5}\selectfont,inner sep=0pt] at (axis cs:1,-0.014559507546) {$ -0.0134 $};
\addplot[ybar,bar width=31.8pt,fill=archMoE,draw=white,line width=0.4pt] coordinates {(2,-0.011582659557)};
\node[anchor=north,font=\sffamily\fontsize{8.5}{9.5}\selectfont,inner sep=0pt] at (axis cs:2,-0.012782659557) {$ -0.0116 $};
\addplot[ybar,bar width=31.8pt,fill=archAttn,draw=white,line width=0.4pt] coordinates {(3,-0.022647080943)};
\node[anchor=north,font=\sffamily\fontsize{8.5}{9.5}\selectfont,inner sep=0pt] at (axis cs:3,-0.023847080943) {$ -0.0226 $};
\addplot[black,dashed,line width=0.7pt] coordinates {(-0.6,0) (3.6,0)};
\node[anchor=north east,font=\sffamily\bfseries\fontsize{9}{10}\selectfont,inner sep=0pt] at (rel axis cs:0.97,0.96) {142M};
\end{axis}
\begin{axis}[stackedbar,at={(0,0in)},anchor=south west,xticklabels={Dense,GDN,MoE,AttnRes}]
\addplot[ybar,bar width=31.8pt,fill=archDense,draw=white,line width=0.4pt] coordinates {(0,-0.006519492716)};
\node[anchor=north,font=\sffamily\fontsize{8.5}{9.5}\selectfont,inner sep=0pt] at (axis cs:0,-0.007719492716) {$ -0.0065 $};
\addplot[ybar,bar width=31.8pt,fill=archGDN,draw=white,line width=0.4pt] coordinates {(1,-0.003654053435)};
\node[anchor=north,font=\sffamily\fontsize{8.5}{9.5}\selectfont,inner sep=0pt] at (axis cs:1,-0.004854053435) {$ -0.0037 $};
\addplot[ybar,bar width=31.8pt,fill=archMoE,draw=white,line width=0.4pt] coordinates {(2,-0.014409922063)};
\node[anchor=north,font=\sffamily\fontsize{8.5}{9.5}\selectfont,inner sep=0pt] at (axis cs:2,-0.015609922063) {$ -0.0144 $};
\addplot[ybar,bar width=31.8pt,fill=archAttn,draw=white,line width=0.4pt] coordinates {(3,-0.011858664453)};
\node[anchor=north,font=\sffamily\fontsize{8.5}{9.5}\selectfont,inner sep=0pt] at (axis cs:3,-0.013058664453) {$ -0.0119 $};
\addplot[black,dashed,line width=0.7pt] coordinates {(-0.6,0) (3.6,0)};
\node[anchor=north east,font=\sffamily\bfseries\fontsize{9}{10}\selectfont,inner sep=0pt] at (rel axis cs:0.97,0.96) {330M};
\end{axis}
\node[rotate=90,font=\sffamily\fontsize{9}{10}\selectfont] at (-0.50in,1.60in) {$\Delta$ final val loss (nats)};
\begin{axis}[at={(3.16in,0in)},anchor=south west,scale only axis,width=2.20in,height=3.20in,
xmin=0.9,xmax=5.3,ymin=3.08,ymax=3.80,xtick={1,2,3,4,5},ytick={3.1,3.2,3.3,3.4,3.5,3.6,3.7,3.8},
tick label style={font=\sffamily\fontsize{8.5}{9.5}\selectfont},
xlabel={training step (K)},ylabel={val loss (nats)},label style={font=\sffamily\fontsize{9}{10}\selectfont},
title={(c) Adding on\'ona (142M Dense)},title style={font=\sffamily\bfseries\fontsize{10}{11}\selectfont,yshift=3pt},
legend style={font=\sffamily\fontsize{8.5}{9.5}\selectfont,draw=none,fill=none,at={(0.98,0.98)},anchor=north east},
legend cell align=left,axis line style={line width=0.7pt},tick style={line width=0.7pt}]
\addplot[gray,dashed,line width=0.7pt,forget plot] coordinates {(2.25,3.08) (2.25,3.80)};
\addplot[curveStd,line width=1.0pt,mark=*,mark size=1.2pt] coordinates {(1.0000,3.708350658417) (1.5000,3.463299512863) (2.0000,3.349214315414) (2.5000,3.274328470230) (3.0000,3.223499059677) (3.5000,3.186846494675) (4.0000,3.153853178024) (4.5000,3.130883216858) (5.0000,3.110360145569) (5.5000,3.092797279358) (6.0000,3.077307462692) (6.5000,3.064822435379)};
\addlegendentry{Standard}
\addplot[curveTwin,line width=1.0pt,mark=*,mark size=1.2pt] coordinates {(1.0000,3.752300024033) (1.5000,3.487006664276) (2.0000,3.352972269058) (2.5000,3.269176721573) (3.0000,3.212995052338) (3.5000,3.166772603989) (4.0000,3.135792016983) (4.5000,3.109363317490) (5.0000,3.087425470352) (5.5000,3.068975448608) (6.0000,3.052747726440) (6.5000,3.039650917053)};
\addlegendentry{on\'ona}
\node[anchor=south west,text=curveTwin,font=\sffamily\fontsize{8.5}{9.5}\selectfont,inner sep=0pt] at (axis cs:1.3,3.66) {on\'ona higher};
\node[text=curveTwin,font=\sffamily\fontsize{8.5}{9.5}\selectfont,inner sep=0pt] at (axis cs:4.4,3.22) {on\'ona lower};
\end{axis}
\end{tikzpicture}}
\endgroup\end{minipage}
\caption{\textbf{\method{} overview and validation loss.} (a) Twin embedding and twin unembedding share two tables. The input uses their average, and the output mixes their softmax distributions. The dashed path shows the input-token router. (b) Final 40-TPP loss of \method{} minus Standard (Table~\ref{tab:main}) at 142M (top) and 330M (bottom); more negative is better. Dashed lines mark Standard. One seed per bar, so no error bars. (c) Early training of 142M Dense; the dashed line marks where the curves cross.}
\label{fig:method}
\vspace{-20pt}
\end{figure}

We make four contributions.
\begin{enumerate}[leftmargin=16pt,topsep=1pt,itemsep=2pt,parsep=0pt]
    \item \textbf{Twin tied tables for embedding and unembedding.} We introduce \method{}, which ties both vocabulary tables to the output head, feeds their average to the unchanged Transformer body, and mixes their two softmax distributions.
    \item \textbf{A gate that reads only the input token.} The mixture weight depends on the current token rather than the hidden state. At serving time, the input tables fold into their average and the gate becomes a lookup of $V$ precomputed values. Exact scoring and greedy decoding still require two vocabulary projections (\S\ref{sec:method} and Appendix~\ref{app:code}).
    \item \textbf{Evaluation across four architectures and three sizes.} We train Dense, GDN hybrid, MoE and AttnRes models at 142M, 330M and 1B with equal token budgets within each size. Eleven of twelve pairs improve validation loss and both external perplexity benchmarks, and ten improve the eight-task accuracy average (Tables~\ref{tab:main}--\ref{tab:downstream}).
    \item \textbf{Controls and ablations.} We compare against an untied head, MoS, input-only gating and a deeper baseline. In a $2\times2$ input--output ablation, neither role of the second table recovers the gain on its own (\S\ref{sec:analysis}).
\end{enumerate}

\section{\Twin{} with Tied Unembeddings}
\label{sec:method}

\gate{} expands the output family of a tied language model by sharing two vocabulary tables between the input and output. Their average feeds a single decoder, whose hidden state is projected through each table to form two softmax distributions. A token-conditioned gate combines these distributions. We first describe the single-table model and its output restriction, then define the twin tables and gate before discussing their expressivity and cost.

\subsection{Preliminaries}
\label{sec:prelim}

Given a token sequence $x_1,\ldots,x_n$ over a vocabulary of size $V$, a decoder-only language model~\citep{vaswani2017attention} with tied embeddings~\citep{press2017using,inan2017tying} computes the next-token distribution in three stages.

\paragraph{Embedding lookup.} The token at position $j$ is mapped to a row of the embedding table $\mE\in\R^{V\times d}$ according to
\begin{equation}
\ve_j=\mE_{x_j}.
\label{eq:lookup}
\end{equation}

The decoder processes the embedding sequence to produce a contextual hidden state $\vh_t$ at each position $t$.

\paragraph{Unembedding (LM head).} The output layer uses the same table to score the vocabulary and predict the next token,
\begin{equation}
\vz_t=\mE\vh_t,\qquad p(x_{t+1}\mid x_{\le t})=\operatorname{softmax}(\vz_t)_{x_{t+1}}.
\label{eq:tied}
\end{equation}
Each row receives gradients from its input and output roles.

\paragraph{Softmax bottleneck.} The matrix of log-probabilities across contexts has rank at most $d+1$~\citep{yang2018breaking}. A \emph{mixture of softmaxes} (MoS) can exceed this bound by combining $K$ distributions from transformed hidden states under one shared table,
\begin{equation}
p(\cdot\mid x_{\le t})=\sum_{k=1}^{K}\pi_k(\vh_t)\cdot\operatorname{softmax}\!\big(\mE\tanh(\mW_k\vh_t)\big),
\label{eq:mos}
\end{equation}
where $\pi_k\ge0$ and $\sum_k\pi_k=1$. This requires $K$ vocabulary projections per position.

\subsection{Architecture of \gate{}}
\label{sec:twin}
\begin{figure}[t]
\centering
\begin{minipage}[t]{0.52\linewidth}
\begin{lstlisting}[style=python,basicstyle=\ttfamily\scriptsize]
class TwinTables(nn.Module):
  def __init__(self, V, d, std=0.02):
    super().__init__()
    self.A = nn.Parameter(std*torch.randn(V, d))
    self.B = nn.Parameter(std*torch.randn(V, d))
    self.w = nn.Linear(d, 1, bias=False)

  def merged(self):  # M: the served input table
    return 0.5 * (self.A + self.B)

  def embed(self, ids):  # train = serve
    return F.embedding(ids, self.merged())

  def gate_logit(self, x0):  # x0 = M[ids]
    # reads the input token only
    return self.w(F.rms_norm(x0, x0.shape[-1:]))
\end{lstlisting}
\end{minipage}\hfill
\begin{minipage}[t]{0.47\linewidth}
\begin{lstlisting}[style=python,basicstyle=\ttfamily\scriptsize]
  # TwinTables, continued
  def log_probs(self, h, x0):  # h: [B,T,d]
    z = self.gate_logit(x0).float()
    la = (h @ self.A.T).float()
    lb = (h @ self.B.T).float()
    la = F.log_softmax(la, -1)
    lb = F.log_softmax(lb, -1)
    # log g, log(1-g): finite if g saturates
    return torch.logaddexp(
      la + F.logsigmoid(z),
      lb + F.logsigmoid(-z))

# training step (body unchanged)
x0 = twin.embed(ids)
h = final_norm(body(x0))
logp = twin.log_probs(h, x0)
loss = F.nll_loss(logp.flatten(0, 1),
                  tgt.flatten())
\end{lstlisting}
\end{minipage}
\vspace{-0.08in}
\caption{\textbf{\method{} pseudocode.} The body reads averaged embeddings, and the head combines two tied softmax distributions using an input-token gate. Appendix~\ref{app:code} provides implementation and serving details.}
\label{fig:code}
\vspace{-0.12in}
\end{figure}

\gate{} has three components. \textbf{Twin embedding} averages two input representations, \textbf{twin unembedding} produces two output distributions from the same hidden state, and a \textbf{token gate} learns how to combine them (Figure~\ref{fig:method}(a)). The input and output components share the learned tables $\mA,\mB\in\R^{V\times d}$.

\paragraph{Twin embedding.} Each input representation is the arithmetic mean of the corresponding rows in the two tables,
\begin{equation}
\mM=\tfrac12(\mA+\mB),\qquad \ve_j=\mM_{x_j}.
\label{eq:twin-input}
\end{equation}
A fixed average gives both tables a symmetric input role without an input router, and it keeps the representation width, so the decoder is untouched.

\paragraph{Twin unembedding.} Each table scores the same $\vh_t$. We normalize the two logit vectors separately and then combine their probabilities,
\begin{equation}
p(\cdot\mid x_{\le t})=g_t\cdot\operatorname{softmax}(\mA\vh_t)
+(1-g_t)\cdot\operatorname{softmax}(\mB\vh_t).
\label{eq:twin}
\end{equation}
The input average is fixed; only the output mixture weight is learned. Normalizing the two logit vectors separately is what lets the head go beyond a single softmax. Mixing before the softmax, as in $\operatorname{softmax}(\mM\vh_t)$, would keep the single-softmax restriction.

\paragraph{Token gate and training.} A vector $\vw\in\R^d$ maps the input token's averaged embedding to the mixture weight,
\begin{equation}
z_t=\vw^\top\operatorname{rmsnorm}(\mM_{x_t}),\qquad g_t=\sigma(z_t).
\label{eq:gate}
\end{equation}
Because the gate reads only the token, each token type learns one mixture weight, which can be precomputed for serving. The gate itself ignores context, but both component distributions depend on it through $\vh_t$. Tables, gate and decoder are trained jointly on the next-token negative log-likelihood. Figure~\ref{fig:code} gives a log-space implementation, and Appendix~\ref{app:code} covers the gradient paths and token routing.

\subsection{Properties and computational cost}
\label{sec:twin-design}

\paragraph{Tying and expressivity.} Both scoring tables also serve as input tables. Mixing their distributions \emph{after} the softmax can exceed the single-softmax $d+1$ rank bound; Appendix~\ref{app:rank} gives an explicit construction and contrasts it with mixing the tables before the softmax. The tied baseline is a special case, since $\mA=\mB=\mE$ recovers it for every gate value. Our experiments do not start from this construction; the two tables are initialized independently (Appendix~\ref{app:setup}).

\paragraph{Relation to MoS.} MoS combines distributions from transformed hidden states under one table, whereas \gate{} uses two tables with the same hidden state. The gate in \gate{} depends on token identity, while MoS uses a contextual prior. Sections~\ref{sec:output-comparison} and~\ref{sec:controls} compare the two designs empirically.

\paragraph{Cost.} The second table and gate vector add $Vd+d$ trainable parameters, and training evaluates both output components, incurring extra projection FLOPs. At inference, conditional computation can avoid this extra projection, analogous to sparse activation in MoE~\citep{fedus2022switch}. The input tables fold into $\mM$ and gate values are precomputed. For untruncated sampling at temperature 1, we first select a component with probability $g_t$ and then compute only its vocabulary projection and softmax. This produces an exact sample from the mixture, restricting the extra projection cost to training for this generation setting. Exact scoring, greedy decoding, and other temperatures or top-$k$/top-$p$ truncation generally still require both projections. Appendix~\ref{app:code} gives the sampling argument, memory costs and measured throughput.

\section{Experiments}
\label{sec:exp}

\subsection{Experimental setup}
\label{sec:setup}

\paragraph{Models and data.} We train Qwen3-style decoders~\citep{yang2025qwen3} from scratch on ClimbMix~\citep{diao2025climb} at 142M, 330M and 1B scales, using GPT-2 BPE~\citep{radford2019language} and 1,024-token contexts. The four architectures are Dense, GDN hybrid~\citep{yang2025gated}, MoE~\citep{fedus2022switch} and AttnRes~\citep{kimi2025attention}. At 142M and 330M, they share depth and width, with MoE active parameters matched to Dense. The smaller models use a common 7.18B-token training pool, and the 1B models use a disjoint 23.9B-token pool. Training repeats data when the token budget exceeds the pool size.

\paragraph{Training and comparisons.} All experiments, including controls, use a budget of 40 tokens per parameter (TPP) based on the Dense Standard parameter count. Variants at each size therefore receive the same number of training tokens. We use AdamW~\citep{loshchilov2019decoupled} with a warmup--stable--decay schedule~\citep{hu2024minicpm}. \emph{Standard} uses a single tied table. \gate{} retains its decoder body and changes the vocabulary tables and output head. The twin tables are initialized independently, so their average has a lower initial variance than the Standard input table. We compare alternative output layers in Section~\ref{sec:output-comparison} and examine component and resource controls in Section~\ref{sec:analysis}. Appendix~\ref{app:setup} provides model configurations, hyperparameters and initialization details, and Appendix~\ref{app:variants} lists the implementation names.

\subsection{Main results}
\label{sec:main}

We compare \gate{} with the single-table Standard baseline at the final 40-TPP checkpoints. Table~\ref{tab:main} reports validation loss, external perplexity and one-table ablations, and Table~\ref{tab:downstream} reports downstream accuracy. Each arm has one training seed. All 142M and 330M checkpoints are scored on the same 1,024 held-out windows. The 1B runs and the controls in Section~\ref{sec:analysis} use separate evaluation settings, described in Appendices~\ref{app:setup} and~\ref{app:code}; Appendix~\ref{app:legacyresults} records an earlier evaluation round under different settings.

\suppressfloats[t]
\begin{table}[!htbp]
\centering
\small
\caption{\textbf{Pretraining results on ClimbMix at 40 TPP} (seed 0). Loss is measured in nats, and $\Delta$ denotes \gate{} minus Standard. Lower values are better. PPL is LAMBADA last-word or WikiText-2 word perplexity. A/B-only uses the selected table at both input and output. The 142M/330M losses use shared validation windows, while 1B uses separate windows and coarser rounding. Appendix~\ref{app:setup} gives evaluation details.}
\label{tab:main}
\begingroup\footnotesize\setlength{\tabcolsep}{2.4pt}
\begin{tabular}{@{}llccccccccc@{}}
\toprule
 & & \multicolumn{3}{c}{Val loss} & \multicolumn{2}{c}{LAMBADA PPL} & \multicolumn{2}{c}{WikiText PPL} & \multicolumn{2}{c}{One table} \\
\cmidrule(lr){3-5}\cmidrule(lr){6-7}\cmidrule(lr){8-9}\cmidrule(lr){10-11}
Size & Arch. & Std. & \gate{} & $\Delta$ & Std. & \gate{} & Std. & \gate{} & $\mA$ & $\mB$ \\
\midrule
142M & Dense & 2.8142 & \textbf{2.7857} & \textbf{$-$0.0285} & 67.4 & \textbf{63.4} & 63.0 & \textbf{60.7} & 3.17 & 3.11 \\
 & GDN hybrid & 2.7600 & \textbf{2.7467} & \textbf{$-$0.0134} & 59.9 & \textbf{55.4} & 57.9 & \textbf{57.3} & 3.11 & 3.05 \\
 & MoE & 2.7190 & \textbf{2.7074} & \textbf{$-$0.0116} & 50.7 & \textbf{49.6} & 54.6 & \textbf{53.4} & 2.92 & 3.49 \\
 & AttnRes & 2.7791 & \textbf{2.7564} & \textbf{$-$0.0226} & 54.8 & \textbf{52.3} & 58.4 & \textbf{56.3} & 3.15 & 3.08 \\
\addlinespace
330M & Dense & 2.5868 & \textbf{2.5803} & \textbf{$-$0.0065} & 30.6 & \textbf{29.6} & 42.7 & \textbf{42.2} & 2.84 & 2.82 \\
 & GDN hybrid & 2.5440 & \textbf{2.5404} & $-$0.0037 & 25.5 & \textbf{24.3} & 39.7 & \textbf{39.4} & 2.73 & 2.82 \\
 & MoE & 2.5183 & \textbf{2.5039} & \textbf{$-$0.0144} & 25.8 & \textbf{23.5} & 38.3 & \textbf{37.1} & 2.81 & 3.36 \\
 & AttnRes & 2.5533 & \textbf{2.5414} & \textbf{$-$0.0119} & 24.5 & \textbf{24.3} & 38.8 & \textbf{38.3} & 2.79 & 2.77 \\
\addlinespace
1B & Dense
 & 2.4902 & \textbf{2.4811} & $\mathbf{-0.009}$
 & 20.0 & \textbf{19.1}
 & 27.3 & \textbf{26.8}
 & 2.67 & 2.66 \\
 & GDN hybrid
 & 2.4686 & \textbf{2.4629} & $\mathbf{-0.006}$
 & 19.4 & \textbf{18.9}
 & 26.3 & \textbf{26.1}
 & 2.64 & 2.64 \\
 & MoE
 & \textbf{2.4967} & 2.5238 & $+0.027$
 & \textbf{22.6} & 22.8
 & \textbf{27.7} & 28.1
 & 7.24 & 6.43 \\
 & AttnRes
 & 2.4766 & \textbf{2.4673} & $\mathbf{-0.009}$
 & 18.6 & \textbf{17.4}
 & 26.3 & \textbf{25.9}
 & 2.66 & 2.65 \\
\bottomrule\end{tabular}\endgroup

\vspace{-0.1in}
\end{table}

\begin{table}[t]
\centering
\small
\caption{\textbf{Downstream accuracy (\%).} Evaluation is zero-shot except for MMLU, which uses up to five shots. HS and LAMB. abbreviate HellaSwag and LAMBADA. Avg. is the unweighted mean of eight tasks, with the higher paired average in bold. HS and ARC use UTF-8 byte-normalized scores from our custom evaluator. The 1B rows come from a separate evaluation run. Appendix~\ref{app:code} describes the benchmarks and scoring.}
\label{tab:downstream}
\resizebox{\linewidth}{!}{
\begingroup\footnotesize\setlength{\tabcolsep}{4.0pt}
\begin{tabular}{@{}lllccccccccc@{}}
\toprule
 & & & & \multicolumn{2}{c}{ARC} & & & & & & \\
\cmidrule(lr){5-6}
Size & Model & Arm & HS & E & C & PIQA & BoolQ & SciQ & LAMB. & MMLU & Avg. \\
\midrule
142M & Dense & Std. & 36.8 & 52.4 & 28.5 & 67.8 & 56.1 & 85.0 & 27.7 & 25.5 & 47.46 \\
 &  & \gate{} & 38.1 & 54.5 & 30.5 & 68.8 & 58.8 & 83.9 & 29.1 & 27.0 & \textbf{48.84} \\
\addlinespace[1pt]
 & GDN hybrid & Std. & 39.3 & 56.3 & 28.0 & 69.3 & 56.6 & 83.7 & 27.3 & 27.1 & \textbf{48.43} \\
 &  & \gate{} & 39.5 & 54.6 & 28.8 & 70.0 & 53.6 & 84.0 & 30.2 & 25.7 & 48.30 \\
\addlinespace[1pt]
 & MoE & Std. & 40.0 & 55.9 & 30.1 & 70.1 & 57.0 & 84.9 & 30.4 & 24.5 & \textbf{49.11} \\
 &  & \gate{} & 40.8 & 55.8 & 30.2 & 70.9 & 51.7 & 86.2 & 30.8 & 26.1 & 49.06 \\
\addlinespace[1pt]
 & AttnRes & Std. & 38.4 & 55.7 & 28.1 & 68.3 & 59.8 & 84.4 & 29.8 & 26.1 & 48.82 \\
 &  & \gate{} & 39.0 & 56.0 & 30.6 & 69.2 & 57.2 & 85.0 & 29.6 & 26.0 & \textbf{49.07} \\
\midrule
330M & Dense & Std. & 46.5 & 61.2 & 33.9 & 71.3 & 55.3 & 88.6 & 36.3 & 26.0 & 52.39 \\
 &  & \gate{} & 47.1 & 62.5 & 32.6 & 72.0 & 61.2 & 88.5 & 35.6 & 26.7 & \textbf{53.27} \\
\addlinespace[1pt]
 & GDN hybrid & Std. & 49.6 & 64.1 & 33.6 & 72.9 & 51.7 & 88.8 & 36.5 & 25.4 & 52.82 \\
 &  & \gate{} & 49.2 & 61.2 & 32.7 & 72.6 & 57.2 & 87.8 & 38.5 & 28.8 & \textbf{53.49} \\
\addlinespace[1pt]
 & MoE & Std. & 50.9 & 64.6 & 34.0 & 73.3 & 54.0 & 87.5 & 36.7 & 25.5 & 53.31 \\
 &  & \gate{} & 51.7 & 63.6 & 34.6 & 73.9 & 55.2 & 87.3 & 38.8 & 28.1 & \textbf{54.14} \\
\addlinespace[1pt]
 & AttnRes & Std. & 48.3 & 62.5 & 33.4 & 72.8 & 55.6 & 88.7 & 37.0 & 25.3 & 52.94 \\
 &  & \gate{} & 49.3 & 63.2 & 35.0 & 73.0 & 58.0 & 89.5 & 37.5 & 27.4 & \textbf{54.11} \\
\midrule
1B & Dense & Std. & 58.14 & 68.90 & 40.44 & 75.46 & 58.38 & 92.30 & 41.92 & 27.00 & 57.82 \\
 &  & \gate{} & 58.71 & 69.91 & 40.96 & 75.57 & 55.17 & 91.30 & 43.64 & 29.37 & \textbf{58.08} \\
\addlinespace[1pt]
 & GDN hybrid & Std. & 59.90 & 69.19 & 41.72 & 76.77 & 59.97 & 93.00 & 42.95 & 28.91 & 59.05 \\
 &  & \gate{} & 60.06 & 68.39 & 39.93 & 76.12 & 64.04 & 93.20 & 44.23 & 27.17 & \textbf{59.14} \\
\addlinespace[1pt]
 & MoE & Std. & 56.34 & 66.41 & 38.23 & 74.48 & 59.39 & 91.80 & 40.79 & 26.35 & 56.72 \\
 &  & \gate{} & 56.19 & 66.16 & 40.10 & 74.54 & 61.31 & 92.60 & 41.49 & 29.55 & \textbf{57.74} \\
\addlinespace[1pt]
 & AttnRes & Std. & 58.65 & 69.36 & 40.70 & 76.22 & 61.96 & 93.10 & 43.97 & 29.50 & 59.18 \\
 &  & \gate{} & 59.01 & 69.15 & 41.21 & 75.73 & 63.88 & 92.30 & 44.36 & 32.82 & \textbf{59.81} \\
\bottomrule\end{tabular}\endgroup

}
\vspace{-0.1in}
\end{table}

\paragraph{Validation loss.} \method{} has lower final validation loss than Standard in all eight 142M and 330M pairs. At 142M the largest gains are Dense and AttnRes (0.0285 and 0.0226 nats); at 330M it is MoE (0.0144). From 142M to 330M the Dense, GDN hybrid and AttnRes gains shrink, while the MoE gain grows from 0.0116 to 0.0144.

\paragraph{External perplexity (142M and 330M).} We measure zero-shot LAMBADA~\citep{paperno2016lambada} last-word perplexity per word, following lm-evaluation-harness~\citep{gao2024harness}, and WikiText-2~\citep{merity2017pointer} test word perplexity using the served model. Both improve in every pair: LAMBADA by 0.99--9.18\% and WikiText-2 by 0.67--3.62\%. The largest LAMBADA reduction is 330M MoE (25.84 to 23.47) and the largest WikiText-2 reduction is 142M AttnRes (58.44 to 56.32). The direction matches validation loss, but the size of the gain and the ranking of architectures differ between the two test sets.

\paragraph{1B results.} At 1B, \method{} lowers validation loss for Dense (2.4902 to 2.4811), GDN hybrid (2.4686 to 2.4629) and AttnRes (2.4766 to 2.4673), but raises it for MoE (2.4967 to 2.5238). The 1B MoE is also worse on LAMBADA and WikiText-2 (22.6 to 22.8 and 27.7 to 28.1), while the other three architectures improve on both. The gain therefore does not carry over to every architecture at 1B. The 1B runs use their own validation windows, so their absolute losses are not comparable with the smaller models.

\paragraph{One-table ablations.} Using only $\mA$ or only $\mB$ at both input and output raises validation loss in all sixteen 142M and 330M ablations, by 0.188--0.855 nats over the corresponding mixture, so the trained models depend on both tables. Because this ablation changes the input and the head together, it says nothing about their separate contributions; Section~\ref{sec:factorial} tests those one at a time.

\paragraph{Downstream accuracy.} The eight-task average improves in six of the eight 142M and 330M pairs (Table~\ref{tab:downstream}). At 142M, Dense gains 1.38 points (47.46\% to 48.84\%) and AttnRes gains 0.25, while GDN hybrid and MoE decline slightly ($-0.14$ and $-0.05$). All four 330M averages improve, by 0.67 to 1.17 points. The per-task picture is noisier: 142M MoE loses 5.32 points on BoolQ while gaining on six other tasks, and 330M GDN hybrid gains 5.47 points on BoolQ and 3.41 on MMLU but loses on five tasks. Only 330M AttnRes improves on all eight. At 1B, all four averages improve, by 0.09 to 1.02 points; the 1.02 is MoE, despite its worse loss. With one seed per arm, differences of a few tenths of a point are not reliable.

\subsection{Comparison with alternatives}
\label{sec:output-comparison}
We also compare \gate{} with other ways of making the head more expressive: MoS ($K{=}2$)~\citep{yang2018breaking}, Sigsoftmax~\citep{kanai2018sigsoftmax}, its two-component mixture (MoSS), and an untied head. All use the 142M Dense backbone and the 40-TPP recipe (seed 0) of Section~\ref{sec:factorial}, whose Standard and \gate{} runs we reuse. Final checkpoints are evaluated as in Tables~\ref{tab:main} and~\ref{tab:downstream}; implementation and measurement details are in Appendix~\ref{app:code}.

\begin{table}[t]
\centering
\caption{\textbf{Alternative output layers on 142M Dense} (seed 0). $\Delta$ is the loss difference from Standard. Bold marks the lowest loss or PPL.}
\label{tab:output-comparison}
\begingroup\footnotesize
\setlength{\tabcolsep}{4pt}
\renewcommand{\arraystretch}{1.08}
\begin{tabular*}{\linewidth}{@{\extracolsep{\fill}}lrrrrrrr@{}}
\toprule
 & \multicolumn{2}{c}{Validation} & \multicolumn{2}{c}{Perplexity $\downarrow$} & \multicolumn{3}{c}{Training cost} \\
\cmidrule(lr){2-3}\cmidrule(lr){4-5}\cmidrule(lr){6-8}
Method & Loss $\downarrow$ & $\Delta$ & LAMBADA & WikiText-2 & Params (M) & ktok/s & GB \\
\midrule
Standard & 2.8069 & --- & 66.03 & 62.69 & 142.47 & 614 & 19.48 \\
\addlinespace[2pt]
MoS ($K{=}2$) & 2.8239 & $+0.0171$ & 53.29 & 65.07 & 143.65 & 451 & 32.78 \\
Sigsoftmax & 2.7992 & $-0.0076$ & 62.49 & 62.05 & 142.47 & 606 & 19.48 \\
MoSS ($K{=}2$) & 2.7974 & $-0.0094$ & \textbf{46.50} & 61.36 & 143.65 & 428 & 33.35 \\
Untied head & 2.8132 & $+0.0064$ & 71.49 & 64.48 & 181.10 & 616 & 20.25 \\
\midrule
\gate{} & \textbf{2.7868} & $\mathbf{-0.0200}$ & 58.38 & \textbf{61.35} & 181.10 & 422 & 38.42 \\
\bottomrule
\end{tabular*}
\endgroup
\end{table}

\paragraph{Pretraining.} \gate{} has the lowest validation loss (2.7868), 0.0106 nats below MoSS, the next best (Table~\ref{tab:output-comparison}). MoSS wins on LAMBADA perplexity (46.50 versus 58.38), and the two are level on WikiText-2 (61.36 versus 61.35).

\begin{table}[t]
\centering
\caption{\textbf{Downstream accuracy (\%) for alternative output layers.} All results use seed 0 and the evaluation protocol of Table~\ref{tab:downstream}. Bold marks the highest value in each column.}
\label{tab:output-downstream}
\begingroup\footnotesize
\setlength{\tabcolsep}{3pt}
\renewcommand{\arraystretch}{1.08}
\begin{tabular*}{\linewidth}{@{\extracolsep{\fill}}lrrrrrrrrr@{}}
\toprule
 & & \multicolumn{2}{c}{ARC} & & & & & & \\
\cmidrule(lr){3-4}
Method & HS & E & C & PIQA & BoolQ & SciQ & LAMB. & MMLU & Avg. \\
\midrule
Standard & 37.01 & 54.17 & \textbf{30.38} & 68.01 & 51.28 & \textbf{87.20} & 27.98 & 25.82 & 47.73 \\
\addlinespace[2pt]
MoS ($K{=}2$) & 36.44 & 54.04 & 29.61 & 67.63 & 54.01 & 85.70 & 29.79 & 26.93 & 48.02 \\
Sigsoftmax & 36.89 & 53.96 & 29.10 & \textbf{69.37} & \textbf{61.83} & 82.70 & 27.69 & 26.01 & 48.44 \\
MoSS ($K{=}2$) & 37.37 & 53.49 & 28.50 & 68.50 & 53.76 & 81.30 & \textbf{31.28} & 26.76 & 47.62 \\
Untied head & 36.73 & 53.20 & 29.69 & 68.17 & 52.20 & 83.20 & 27.46 & 26.80 & 47.18 \\
\midrule
\gate{} & \textbf{37.99} & \textbf{54.42} & 30.29 & 68.72 & 59.20 & 84.70 & 30.33 & \textbf{27.31} & \textbf{49.12} \\
\bottomrule
\end{tabular*}
\endgroup
\end{table}

\paragraph{Downstream performance.} \gate{} also has the highest eight-task average, 49.12\% (Table~\ref{tab:output-downstream}), 1.39 points above Standard and 0.68 above Sigsoftmax, the next best. No method leads on every task.

\paragraph{Resource tradeoff.} Sigsoftmax improves loss and average accuracy with no extra parameters and near-baseline throughput. \gate{} gains more but costs more than MoSS in parameters and memory (181.10M/38.42GB versus 143.65M/33.35GB), at similar throughput (422 versus 428 ktok/s). All comparisons match training tokens, not parameters or FLOPs, and use one seed with no per-method tuning.

\section{Ablation Analysis}
\label{sec:analysis}
The main results leave three questions open. Is the gain just extra vocabulary capacity, or just extra output computation (Section~\ref{sec:controls})? How much comes from the second table's input role, its output role, and their interaction (Section~\ref{sec:factorial})? And would the same budget be better spent on decoder depth (Section~\ref{sec:depth-control})? Each control group has its own Standard baseline, and comparisons are made within the group.

\subsection{Vocabulary capacity and output mixtures}
\label{sec:controls}

Table~\ref{tab:controls} compares other ways of spending the extra parameters and output computation, to check whether the gain can be had without coupling the second table's input and output roles.

\begin{table}[t]
\centering
\small
\caption{\textbf{Dense controls at 40 TPP.} Extra params gives the added parameter count, Head cost counts vocabulary projections per position, and Serving gives the number of tables after folding. $\Delta$ is validation loss minus Standard, using the two-seed mean at 142M. $^\ddagger$ marks a different GPU layout with the same global batch.}
\label{tab:controls}
\begingroup\footnotesize\setlength{\tabcolsep}{2.6pt}
\begin{tabular}{@{}lcccccc@{}}
\toprule
Model & Tables & Extra params & Head cost & Serving & Val & $\Delta$ \\
\midrule
\multicolumn{7}{@{}l}{\emph{142M Dense, 40 TPP ($\Delta$ against the mean of two Standard seeds, 2.8777 and 2.8693)}} \\
Standard (tied), 2 seeds & 1 & -- & 1$\times$ & 1 table & 2.8735 & -- \\
Untied head$^\ddagger$ & 2 & $Vd$ & 1$\times$ & 2 tables & 2.8821 & $+$0.009 \\
MoS $K{=}2$ (two contexts, one table)$^\ddagger$ & 1 & $2d^2{+}2d$ & 2$\times$ & 1 table & 2.8977 & $+$0.024 \\
Input gate only$^\ddagger$, 2 seeds & 2 & $Vd{+}d$ & 1$\times$ & 2 tables & 2.8641 & $-$0.009 \\
\gate{}$^\ddagger$, 2 seeds & 2 & $Vd{+}d$ & 2$\times$ & 2 tables & \textbf{2.8505} & \textbf{$-$0.023} \\
\midrule
\multicolumn{7}{@{}l}{\emph{330M Dense, 40 tokens per parameter}} \\
Standard, 24 layers & 1 & -- & 1$\times$ & 1 table & 2.6426 & -- \\
Untied head$^\ddagger$ & 2 & $Vd$ & 1$\times$ & 2 tables & 2.6457 & $+$0.003 \\
Standard, 28 layers (approx. FLOP-matched) & 1 & $4$ layers & 1$\times$ & 1 table & \textbf{2.6258} & \textbf{$-$0.017} \\
\gate{}$^\ddagger$ & 2 & $Vd{+}d$ & 2$\times$ & 2 tables & 2.6329 & $-$0.010 \\
\bottomrule\end{tabular}\endgroup

\vspace{-0.1in}
\end{table}

\paragraph{Capacity and output mixtures.} An untied head adds the same $V\times d$ matrix but raises validation loss by 0.009 nats at 142M and 0.003 at 330M. MoS ($K{=}2$) matches the number of vocabulary projections but raises loss by 0.024 nats at 142M. Neither extra vocabulary capacity nor a second softmax on its own reproduces the gain under this recipe. Section~\ref{sec:output-comparison} covers further output layers in a separate matched group.

\paragraph{Input-side routing.} A learned input mixture with an averaged single-table head lowers loss by 0.009 nats at 142M, against 0.023 for \gate{}, both relative to the two-seed Standard mean. That is less than half the gain, but this control changes both the input rule and the head, so the remaining gap cannot be pinned on output gating alone.

\subsection{Interaction between input and output roles}
\label{sec:factorial}
\label{sec:interaction}

\paragraph{Matched factorial design.}
To separate the input and output roles, we train a $2\times2$ grid of 142M Dense models. The input is either $\mA$ or the fixed average $(\mA+\mB)/2$, and the output is either the $\mA$ softmax or the gated two-table mixture. Each cell is trained from scratch, so it adapts to its configuration, unlike the A-only/B-only ablations of a trained checkpoint. All four cells share seed 0, initial body and $\mA$ weights, data order, optimizer schedule, token budget and 1,024 validation windows (Appendix~\ref{app:code}). The input-only cell uses a fixed average, not the learned input gate of the control above.

\begin{table}[t]
\centering
\caption{\textbf{Input--output ablation at 142M Dense.} Final validation loss for seed 0, with the difference from this grid's baseline in parentheses. Lower is better.}
\label{tab:factorial}
\setlength{\tabcolsep}{16pt}
\renewcommand{\arraystretch}{1.35}
\begin{tabular}{lcc}
\toprule
 & Single unembedding & Twin unembedding \\
\midrule
Single embedding & \shortstack{2.80685\\$(+0.00000)$} & \shortstack{2.81401\\$(+0.00716)$} \\
Twin embedding & \shortstack{2.79956\\$(-0.00729)$} & \shortstack{\textbf{2.78683}\\$(-0.02002)$} \\
\bottomrule
\end{tabular}
\end{table}

\paragraph{Joint contribution and interaction.}
Input averaging alone lowers loss by 0.00729 nats; the output mixture alone raises it by 0.00716 (Table~\ref{tab:factorial}). Together they lower it by 0.02002. All three two-table cells have the same $2Vd$ vocabulary parameters, and the mixture adds only a $d$-parameter gate, so the extra vocabulary matrix by itself does not explain the joint model's advantage over either single change.

The output mixture helps with averaged inputs and hurts with single-table inputs. Writing $L_{ij}$ for the loss, with $i=1$ for twin embeddings and $j=1$ for twin unembeddings, the difference between the two effects is
\begin{equation}
S=(L_{10}-L_{11})-(L_{00}-L_{01})=0.01990\ \text{nats}.
\label{eq:factorial-interaction}
\end{equation}
The interaction is positive: the joint reduction exceeds the sum of the two single-change effects, so in this grid the two roles are complementary.

\subsection{Depth as an alternative use of resources}
\label{sec:depth-control}

To ask whether the same budget would do more inside the decoder, we deepen the 330M Standard from 24 to 28 layers, which roughly matches the added parameters and training FLOPs of \gate{} (Appendix~\ref{app:code}). Loss falls from 2.6426 to 2.6258, 0.0071 below \gate{} at 2.6329 (Table~\ref{tab:controls}). In this comparison, depth is the better use of the budget. We have not run the corresponding 1B control.

\section{Related Work}
\label{sec:related}

\paragraph{Tied embeddings.} Sharing the input and output tables~\citep{press2017using,inan2017tying} is standard in small and mid-sized models~\citep{liu2024mobilellm,yang2025qwen3}. Output gradients largely shape the shared matrix early in training~\citep{lopardo2026weight}. The fully untied 400M control of \citet{batley2026leviathan} underperforms its tied baseline, consistent with our untying controls. Over-tokenized Transformers~\citep{huang2025overtokenized} increase input vocabulary capacity through extra tables while retaining a single output table. \Twin{} instead use two tables over the same vocabulary and tie both to the output.

\paragraph{The softmax bottleneck.} \citet{yang2018breaking} established the rank bound for a single-softmax head and proposed MoS to overcome it. Subsequent work reduced the cost of richer output layers with Mixtape~\citep{yang2019mixtape} and changed the output activation without adding parameters with Sigsoftmax~\citep{kanai2018sigsoftmax}. Studies of the bottleneck link its effects to multi-mode next-token distributions~\citep{chang2022softmax} and the saturation of small models~\citep{godey2024why}. MoS and \gate{} increase output expressiveness in different ways. MoS varies the context under one table, while \gate{} varies the table under one context and uses a token-dependent prior. Our MoS controls compare the two directly (Tables~\ref{tab:controls} and~\ref{tab:output-comparison}).

\paragraph{Token-conditioned routing.} Hash layers~\citep{roller2021hash} route tokens to experts by identity rather than by a learned function of context. They are competitive with learned routing in the reported sparse-model experiments. \gate{} also uses token identity, but learns soft output-mixture weights rather than fixed hash assignments. Each previous-token type learns how much the two tables contribute to the next-token distribution.

\paragraph{Multiple embeddings per token.} Multi-sense embeddings~\citep{neelakantan2014efficient} and dynamic meta-embeddings~\citep{kiela2018dynamic} keep several input vectors per word and use context to select or attend among them. GloVe combines its two tables by summation at serving time~\citep{pennington2014glove}, while multi-embedding recommenders retain several tables per feature during serving~\citep{guo2024embedding}. \Twin{} also retain two vectors per token. Their distinguishing feature is that both tables are tied to the head, where their output distributions are mixed.

\section{Conclusion}
\label{sec:conclusion}

\method{} shares two vocabulary tables between input and output, feeds their average to the decoder, and combines their softmax distributions with a gate conditioned only on the input token. At 142M and 330M, it lowers validation loss and both external perplexities for all four architectures. The downstream average improves in six of eight pairs, with small declines for 142M GDN hybrid and MoE. At 1B, Dense, GDN hybrid and AttnRes improve, while MoE has worse loss despite a higher downstream average. In the ablations, neither role of the second table recovers the gain on its own, and neither untying nor MoS reproduces it under the tested recipe. The gain costs an extra vocabulary projection, and at 330M a deeper Standard model does better with a similar budget. We have not run multi-seed experiments or a resource-matched comparison at 1B; Appendix~\ref{app:limitations} lists the remaining limitations.

\bibliography{refs}

\begin{thebibliography}{39}
\providecommand{\natexlab}[1]{#1}
\providecommand{\url}[1]{\texttt{#1}}
\expandafter\ifx\csname urlstyle\endcsname\relax
  \providecommand{\doi}[1]{doi: #1}\else
  \providecommand{\doi}{doi: \begingroup \urlstyle{rm}\Url}\fi

\bibitem[Ainslie et~al.(2023)Ainslie, Lee-Thorp, de~Jong, Zemlyanskiy, Lebr{\'o}n, and Sanghai]{ainslie2023gqa}
Joshua Ainslie, James Lee-Thorp, Michiel de~Jong, Yury Zemlyanskiy, Federico Lebr{\'o}n, and Sumit Sanghai.
\newblock {GQA}: Training generalized multi-query transformer models from multi-head checkpoints.
\newblock In \emph{Proceedings of the 2023 Conference on Empirical Methods in Natural Language Processing}, pp.\  4895--4901. Association for Computational Linguistics, 2023.
\newblock \doi{10.18653/v1/2023.emnlp-main.298}.

\bibitem[Batley \& Saha(2026)Batley and Saha]{batley2026leviathan}
Reza~T. Batley and Sourav Saha.
\newblock Leviathan: Decoupling input and output representations in language models.
\newblock \emph{arXiv preprint arXiv:2601.22040}, 2026.

\bibitem[Bisk et~al.(2020)Bisk, Zellers, Le~Bras, Gao, and Choi]{bisk2020piqa}
Yonatan Bisk, Rowan Zellers, Ronan Le~Bras, Jianfeng Gao, and Yejin Choi.
\newblock {PIQA}: Reasoning about physical commonsense in natural language.
\newblock In \emph{Proceedings of the AAAI Conference on Artificial Intelligence}, volume~34, pp.\  7432--7439, 2020.
\newblock \doi{10.1609/aaai.v34i05.6239}.

\bibitem[Chang \& McCallum(2022)Chang and McCallum]{chang2022softmax}
Haw-Shiuan Chang and Andrew McCallum.
\newblock Softmax bottleneck makes language models unable to represent multi-mode word distributions.
\newblock In \emph{Proceedings of the 60th Annual Meeting of the Association for Computational Linguistics (ACL)}, 2022.

\bibitem[Clark et~al.(2019)Clark, Lee, Chang, Kwiatkowski, Collins, and Toutanova]{clark2019boolq}
Christopher Clark, Kenton Lee, Ming-Wei Chang, Tom Kwiatkowski, Michael Collins, and Kristina Toutanova.
\newblock {BoolQ}: Exploring the surprising difficulty of natural yes/no questions.
\newblock In \emph{Proceedings of the 2019 Conference of the North {A}merican Chapter of the Association for Computational Linguistics: Human Language Technologies, Volume 1 (Long and Short Papers)}, pp.\  2924--2936. Association for Computational Linguistics, 2019.
\newblock \doi{10.18653/v1/N19-1300}.

\bibitem[Clark et~al.(2018)Clark, Cowhey, Etzioni, Khot, Sabharwal, Schoenick, and Tafjord]{clark2018think}
Peter Clark, Isaac Cowhey, Oren Etzioni, Tushar Khot, Ashish Sabharwal, Carissa Schoenick, and Oyvind Tafjord.
\newblock Think you have solved question answering? {T}ry {ARC}, the {AI2} reasoning challenge.
\newblock \emph{arXiv preprint arXiv:1803.05457}, 2018.

\bibitem[Diao et~al.(2025)Diao, Yang, Fu, Dong, Su, Kliegl, Chen, Belcak, Suhara, Yin, Patwary, Lin, Kautz, and Molchanov]{diao2025climb}
Shizhe Diao, Yu~Yang, Yonggan Fu, Xin Dong, Dan Su, Markus Kliegl, Zijia Chen, Peter Belcak, Yoshi Suhara, Hongxu Yin, Mostofa Patwary, Yingyan Lin, Jan Kautz, and Pavlo Molchanov.
\newblock {Nemotron-CLIMB}: {CL}ustering-based iterative data mixture bootstrapping for language model pre-training.
\newblock \emph{arXiv preprint arXiv:2504.13161}, 2025.

\bibitem[Fedus et~al.(2022)Fedus, Zoph, and Shazeer]{fedus2022switch}
William Fedus, Barret Zoph, and Noam Shazeer.
\newblock Switch transformers: Scaling to trillion parameter models with simple and efficient sparsity.
\newblock \emph{JMLR}, 2022.

\bibitem[Gao et~al.(2024)Gao, Tow, Abbasi, Biderman, Black, DiPofi, Foster, Golding, Hsu, Le~Noac'h, Li, McDonell, Muennighoff, Ociepa, Phang, Reynolds, Schoelkopf, Skowron, Sutawika, Tang, Thite, Wang, Wang, and Zou]{gao2024harness}
Leo Gao, Jonathan Tow, Baber Abbasi, Stella Biderman, Sid Black, Anthony DiPofi, Charles Foster, Laurence Golding, Jeffrey Hsu, Alain Le~Noac'h, Haonan Li, Kyle McDonell, Niklas Muennighoff, Chris Ociepa, Jason Phang, Laria Reynolds, Hailey Schoelkopf, Aviya Skowron, Lintang Sutawika, Eric Tang, Anish Thite, Ben Wang, Kevin Wang, and Andy Zou.
\newblock The language model evaluation harness.
\newblock Zenodo, 2024.
\newblock URL \url{https://doi.org/10.5281/zenodo.12608602}.
\newblock Version v0.4.3.

\bibitem[Godey et~al.(2024)Godey, de~la Clergerie, and Sagot]{godey2024why}
Nathan Godey, {\'E}ric de~la Clergerie, and Beno{\^\i}t Sagot.
\newblock Why do small language models underperform? {S}tudying language model saturation via the softmax bottleneck.
\newblock \emph{arXiv preprint arXiv:2404.07647}, 2024.

\bibitem[Guo et~al.(2024)Guo, Pan, Wang, Chen, Jiang, and Long]{guo2024embedding}
Xingzhuo Guo, Junwei Pan, Ximei Wang, Baixu Chen, Jie Jiang, and Mingsheng Long.
\newblock On the embedding collapse when scaling up recommendation models.
\newblock In \emph{International Conference on Machine Learning}, 2024.

\bibitem[Hendrycks et~al.(2021)Hendrycks, Burns, Basart, Zou, Mazeika, Song, and Steinhardt]{hendrycks2021measuring}
Dan Hendrycks, Collin Burns, Steven Basart, Andy Zou, Mantas Mazeika, Dawn Song, and Jacob Steinhardt.
\newblock Measuring massive multitask language understanding.
\newblock In \emph{International Conference on Learning Representations}, 2021.
\newblock arXiv:2009.03300.

\bibitem[Henry et~al.(2020)Henry, Dachapally, Pawar, and Chen]{henry2020querykey}
Alex Henry, Prudhvi~Raj Dachapally, Shubham~Shantaram Pawar, and Yuxuan Chen.
\newblock Query-key normalization for transformers.
\newblock In \emph{Findings of the Association for Computational Linguistics: {EMNLP} 2020}, pp.\  4246--4253. Association for Computational Linguistics, 2020.
\newblock \doi{10.18653/v1/2020.findings-emnlp.379}.

\bibitem[Hu et~al.(2024)Hu, Tu, Han, He, Cui, Long, Zheng, Fang, Huang, Zhao, Zhang, Thai, Zhang, Wang, Yao, Zhao, Zhou, Cai, Zhai, Ding, Jia, Zeng, Li, Liu, and Sun]{hu2024minicpm}
Shengding Hu, Yuge Tu, Xu~Han, Chaoqun He, Ganqu Cui, Xiang Long, Zhi Zheng, Yewei Fang, Yuxiang Huang, Weilin Zhao, Xinrong Zhang, Zheng~Leng Thai, Kaihuo Zhang, Chongyi Wang, Yuan Yao, Chenyang Zhao, Jie Zhou, Jie Cai, Zhongwu Zhai, Ning Ding, Chao Jia, Guoyang Zeng, Dahai Li, Zhiyuan Liu, and Maosong Sun.
\newblock {MiniCPM}: Unveiling the potential of small language models with scalable training strategies.
\newblock In \emph{Conference on Language Modeling (COLM)}, 2024.
\newblock arXiv:2404.06395.

\bibitem[Huang et~al.(2025)Huang, Zhu, Wu, Zeng, Wang, Min, and Zhou]{huang2025overtokenized}
Hongzhi Huang, Defa Zhu, Banggu Wu, Yutao Zeng, Ya~Wang, Qiyang Min, and Xun Zhou.
\newblock Over-tokenized transformer: Vocabulary is generally worth scaling.
\newblock In \emph{International Conference on Machine Learning}, 2025.
\newblock arXiv:2501.16975.

\bibitem[Inan et~al.(2017)Inan, Khosravi, and Socher]{inan2017tying}
Hakan Inan, Khashayar Khosravi, and Richard Socher.
\newblock Tying word vectors and word classifiers: A loss framework for language modeling.
\newblock In \emph{International Conference on Learning Representations}, 2017.

\bibitem[Kanai et~al.(2018)Kanai, Fujiwara, Yamanaka, and Adachi]{kanai2018sigsoftmax}
Sekitoshi Kanai, Yasuhiro Fujiwara, Yuki Yamanaka, and Shuichi Adachi.
\newblock Sigsoftmax: Reanalysis of the softmax bottleneck.
\newblock In \emph{Advances in Neural Information Processing Systems (NeurIPS)}, 2018.

\bibitem[Kiela et~al.(2018)Kiela, Wang, and Cho]{kiela2018dynamic}
Douwe Kiela, Changhan Wang, and Kyunghyun Cho.
\newblock Dynamic meta-embeddings for improved sentence representations.
\newblock In \emph{Proceedings of the 2018 Conference on Empirical Methods in Natural Language Processing (EMNLP)}, 2018.

\bibitem[{Kimi Team}(2026)]{kimi2025attention}
{Kimi Team}.
\newblock Attention residuals.
\newblock \emph{arXiv preprint arXiv:2603.15031}, 2026.

\bibitem[Liu et~al.(2024)Liu, Zhao, Iandola, Lai, Tian, Fedorov, Xiong, Chang, Shi, Krishnamoorthi, Lai, and Chandra]{liu2024mobilellm}
Zechun Liu, Changsheng Zhao, Forrest Iandola, Chen Lai, Yuandong Tian, Igor Fedorov, Yunyang Xiong, Ernie Chang, Yangyang Shi, Raghuraman Krishnamoorthi, Liangzhen Lai, and Vikas Chandra.
\newblock {MobileLLM}: Optimizing sub-billion parameter language models for on-device use cases.
\newblock In \emph{International Conference on Machine Learning}, 2024.
\newblock arXiv:2402.14905.

\bibitem[Lopardo et~al.(2026)Lopardo, Harish, Arnett, and Gupta]{lopardo2026weight}
Antonio Lopardo, Avyukth Harish, Catherine Arnett, and Akshat Gupta.
\newblock Weight tying biases token embeddings towards the output space.
\newblock In Maria Liakata, Viviane~P. Moreira, Jiajun Zhang, and David Jurgens (eds.), \emph{Findings of the {A}ssociation for {C}omputational {L}inguistics: {ACL} 2026}, pp.\  40795--40810, San Diego, California, United States, July 2026. Association for Computational Linguistics.
\newblock ISBN 979-8-89176-395-1.
\newblock \doi{10.18653/v1/2026.findings-acl.2027}.
\newblock URL \url{https://aclanthology.org/2026.findings-acl.2027/}.

\bibitem[Loshchilov \& Hutter(2019)Loshchilov and Hutter]{loshchilov2019decoupled}
Ilya Loshchilov and Frank Hutter.
\newblock Decoupled weight decay regularization.
\newblock In \emph{International Conference on Learning Representations}, 2019.
\newblock arXiv:1711.05101.

\bibitem[Merity et~al.(2017)Merity, Xiong, Bradbury, and Socher]{merity2017pointer}
Stephen Merity, Caiming Xiong, James Bradbury, and Richard Socher.
\newblock Pointer sentinel mixture models.
\newblock In \emph{International Conference on Learning Representations}, 2017.

\bibitem[Neelakantan et~al.(2014)Neelakantan, Shankar, Passos, and McCallum]{neelakantan2014efficient}
Arvind Neelakantan, Jeevan Shankar, Alexandre Passos, and Andrew McCallum.
\newblock Efficient non-parametric estimation of multiple embeddings per word in vector space.
\newblock In \emph{Proceedings of the 2014 Conference on Empirical Methods in Natural Language Processing (EMNLP)}, 2014.

\bibitem[Paperno et~al.(2016)Paperno, Kruszewski, Lazaridou, Pham, Bernardi, Pezzelle, Baroni, Boleda, and Fern{\'a}ndez]{paperno2016lambada}
Denis Paperno, Germ{\'a}n Kruszewski, Angeliki Lazaridou, Ngoc~Quan Pham, Raffaella Bernardi, Sandro Pezzelle, Marco Baroni, Gemma Boleda, and Raquel Fern{\'a}ndez.
\newblock The {LAMBADA} dataset: Word prediction requiring a broad discourse context.
\newblock In \emph{Proceedings of the 54th Annual Meeting of the Association for Computational Linguistics (Volume 1: Long Papers)}, pp.\  1525--1534. Association for Computational Linguistics, 2016.
\newblock \doi{10.18653/v1/P16-1144}.

\bibitem[Pennington et~al.(2014)Pennington, Socher, and Manning]{pennington2014glove}
Jeffrey Pennington, Richard Socher, and Christopher~D. Manning.
\newblock {GloVe}: Global vectors for word representation.
\newblock In \emph{Proceedings of the 2014 Conference on Empirical Methods in Natural Language Processing (EMNLP)}, 2014.

\bibitem[Press \& Wolf(2017)Press and Wolf]{press2017using}
Ofir Press and Lior Wolf.
\newblock Using the output embedding to improve language models.
\newblock In \emph{Conference of the European Chapter of the Association for Computational Linguistics}, 2017.

\bibitem[{Qwen Team}(2025)]{yang2025qwen3}
{Qwen Team}.
\newblock {Qwen3} technical report.
\newblock \emph{arXiv preprint arXiv:2505.09388}, 2025.

\bibitem[Radford et~al.(2019)Radford, Wu, Child, Luan, Amodei, and Sutskever]{radford2019language}
Alec Radford, Jeffrey Wu, Rewon Child, David Luan, Dario Amodei, and Ilya Sutskever.
\newblock Language models are unsupervised multitask learners.
\newblock Technical report, OpenAI, 2019.
\newblock URL \url{https://cdn.openai.com/better-language-models/language_models_are_unsupervised_multitask_learners.pdf}.

\bibitem[Roller et~al.(2021)Roller, Sukhbaatar, Szlam, and Weston]{roller2021hash}
Stephen Roller, Sainbayar Sukhbaatar, Arthur Szlam, and Jason Weston.
\newblock Hash layers for large sparse models.
\newblock In \emph{Advances in Neural Information Processing Systems}, 2021.

\bibitem[Shazeer(2020)]{shazeer2020glu}
Noam Shazeer.
\newblock {GLU} variants improve transformer.
\newblock \emph{arXiv preprint arXiv:2002.05202}, 2020.

\bibitem[Su et~al.(2024)Su, Ahmed, Lu, Pan, Bo, and Liu]{su2024roformer}
Jianlin Su, Murtadha Ahmed, Yu~Lu, Shengfeng Pan, Wen Bo, and Yunfeng Liu.
\newblock {RoFormer}: Enhanced transformer with rotary position embedding.
\newblock \emph{Neurocomputing}, 568:\penalty0 127063, 2024.
\newblock \doi{10.1016/j.neucom.2023.127063}.

\bibitem[Vaswani et~al.(2017)Vaswani, Shazeer, Parmar, Uszkoreit, Jones, Gomez, Kaiser, and Polosukhin]{vaswani2017attention}
Ashish Vaswani, Noam Shazeer, Niki Parmar, Jakob Uszkoreit, Llion Jones, Aidan~N. Gomez, {\L}ukasz Kaiser, and Illia Polosukhin.
\newblock Attention is all you need.
\newblock In \emph{Advances in Neural Information Processing Systems}, volume~30, 2017.

\bibitem[Welbl et~al.(2017)Welbl, Liu, and Gardner]{welbl2017crowdsourcing}
Johannes Welbl, Nelson~F. Liu, and Matt Gardner.
\newblock Crowdsourcing multiple choice science questions.
\newblock In \emph{Proceedings of the 3rd Workshop on Noisy User-generated Text}, 2017.

\bibitem[Yang et~al.(2025)Yang, Kautz, and Hatamizadeh]{yang2025gated}
Songlin Yang, Jan Kautz, and Ali Hatamizadeh.
\newblock Gated delta networks: Improving mamba2 with delta rule.
\newblock In \emph{ICLR}, 2025.

\bibitem[Yang et~al.(2018)Yang, Dai, Salakhutdinov, and Cohen]{yang2018breaking}
Zhilin Yang, Zihang Dai, Ruslan Salakhutdinov, and William~W. Cohen.
\newblock Breaking the softmax bottleneck: A high-rank {RNN} language model.
\newblock In \emph{International Conference on Learning Representations (ICLR)}, 2018.

\bibitem[Yang et~al.(2019)Yang, Luong, Salakhutdinov, and Le]{yang2019mixtape}
Zhilin Yang, Thang Luong, Ruslan Salakhutdinov, and Quoc~V. Le.
\newblock Mixtape: Breaking the softmax bottleneck efficiently.
\newblock In \emph{Advances in Neural Information Processing Systems (NeurIPS)}, 2019.

\bibitem[Zellers et~al.(2019)Zellers, Holtzman, Bisk, Farhadi, and Choi]{zellers2019hellaswag}
Rowan Zellers, Ari Holtzman, Yonatan Bisk, Ali Farhadi, and Yejin Choi.
\newblock {HellaSwag}: Can a machine really finish your sentence?
\newblock In \emph{Proceedings of the 57th Annual Meeting of the Association for Computational Linguistics}, pp.\  4791--4800. Association for Computational Linguistics, 2019.
\newblock \doi{10.18653/v1/P19-1472}.

\bibitem[Zhang \& Sennrich(2019)Zhang and Sennrich]{zhang2019root}
Biao Zhang and Rico Sennrich.
\newblock Root mean square layer normalization.
\newblock In \emph{Advances in Neural Information Processing Systems}, volume~32, 2019.

\end{thebibliography}
\bibliographystyle{iclr2027_conference}

\newpage
\appendix
\newsavebox{\twdAppBox}
\newcommand{\twdFit}[1]{\sbox{\twdAppBox}{#1}\ifdim\wd\twdAppBox>\linewidth\resizebox{\linewidth}{!}{\usebox{\twdAppBox}}\else\usebox{\twdAppBox}\fi}

\section{The rank view of the mixture}
\label{app:rank}

\paragraph{Single-softmax bound.}
Let $\mH\in\R^{C\times d}$ contain the final hidden states for $C$ contexts, and let $\mL\in\R^{C\times V}$ contain their log-probability vectors. For a fixed vocabulary table $\mE$, a bias-free single-softmax head gives
\begin{equation}
\mL=\mH\mE^\top-\boldsymbol{\ell}\mathbf{1}^\top,
\qquad \ell_c=\log\sum_{v=1}^{V}\exp((\mH\mE^\top)_{cv}).
\end{equation}
Thus $\operatorname{rank}(\mL)\le\min(C,V,d+1)$ by rank subadditivity~\citep{yang2018breaking}.

\paragraph{A constructive counterexample for the two-table mixture.}
Nonlinearity of $\log p$ alone does not prove a larger rank, so we give an explicit instance. Take the head in Eq.~\ref{eq:twin} with $d=1$, $V=C=3$, and
\begin{equation}
\mA=\begin{pmatrix}0\\1\\2\end{pmatrix},\qquad
\mB=\begin{pmatrix}0\\0\\0\end{pmatrix},\qquad
\mH=\begin{pmatrix}\log 2\\0\\-\log 2\end{pmatrix},\qquad g_c=\tfrac12.
\end{equation}
The constant gate is reachable in our parameterization, since $\vw=0$ gives $g_c=\sigma(0)=1/2$ for every token. We evaluate the head at the displayed hidden states, for which $\operatorname{softmax}(\mB h)$ is uniform. The mixture probability matrix is
\begin{equation}
\mP=\frac1{42}\begin{pmatrix}10&13&19\\14&14&14\\19&13&10\end{pmatrix}.
\end{equation}
Taking the logarithm entrywise, write $a=\log(10/42)$, $b=\log(13/42)$, $c=\log(19/42)$ and $u=\log(1/3)$. Then
\begin{equation}
\mL=\begin{pmatrix}a&b&c\\u&u&u\\c&b&a\end{pmatrix},\qquad
\det\mL=u(a-c)(a+c-2b)
=\log(1/3)\log(10/19)\log(190/169)\ne0.
\end{equation}
So $\operatorname{rank}(\mL)=3>d+1=2$: the two-table mixture can exceed the single-softmax bound even with a constant gate. This says nothing about whether trained models do so, or whether rank explains the measured gains; with $\mA=\mB$ the head is again a single softmax.

\paragraph{Mixing tables before softmax with fixed or varying gates.}
For a fixed gate $g$, let $\mE_g=g\mA+(1-g)\mB$. The pre-softmax mixture $\operatorname{softmax}(\mE_g\vh)$ is a single-softmax head and keeps the $d+1$ bound. With a token-dependent gate the same argument applies within each set of contexts that end in the same token, since those share a gate value, but it gives no global $d+1$ bound across contexts with different gate values.

More generally, let $\mD=\operatorname{diag}(g_1,\ldots,g_C)$. The pre-softmax mixture has stacked logits
\begin{equation}
\mZ=\mD\mH\mA^\top+(\mI-\mD)\mH\mB^\top.
\end{equation}
Each summand has rank at most $d$, so after subtracting the row log-normalizers, $\operatorname{rank}(\mL)\le\min(C,V,2d+1)$. Whether this bound is attained is a separate question; the point is only that the fixed-gate argument cannot rule out a global rank above $d+1$ once the gate varies by token. For the post-softmax mixture, the construction above settles the question directly.

\section{Implementation notes}
\label{app:code}

\paragraph{The head in code.} The following implementation uses $\mathtt{x0}$ for the input row $\mM_{x_t}$ and $\mathtt{h}$ for the final hidden state.
\begin{verbatim}
z  = w(rms_norm(x0))     # [B,T,1] gate logit, token only
la = log_softmax(h @ A.T)    # fp32
lb = log_softmax(h @ B.T)
logp = logaddexp(la + logsigmoid(z), lb + logsigmoid(-z))
loss = cross_entropy(logp, targets)  # logp is normalized
\end{verbatim}
The terms $\operatorname{logsigmoid}(z)$ and $\operatorname{logsigmoid}(-z)$ give $\log g$ and $\log(1-g)$ and stay finite when the gate saturates; an earlier version that took $\log\sigma(z)$ directly diverged once $\sigma(z)$ underflowed. Because $\log p$ is already normalized, the log-softmax inside the cross-entropy is a no-op. $\mathtt{x0}$ is not detached, so gate gradients reach the input rows through RMSNorm.

\paragraph{Serving.} Because the gate depends only on the token, we precompute $\vg=\sigma(\operatorname{rmsnorm}(\mM)\vw)\in[0,1]^V$ and fold the input tables into $\mM$. Exact scoring and greedy decoding still compute both projections $\mA\vh_t$ and $\mB\vh_t$ and both log-softmaxes at each position. Sampling at temperature 1 without truncation is cheaper: draw $c\sim\mathrm{Bernoulli}(g_{x_t})$, then sample from $\operatorname{softmax}(\mA\vh_t)$ or $\operatorname{softmax}(\mB\vh_t)$ accordingly. This is an exact sample from the mixture with one projection. Other temperatures and top-$k$ or top-$p$ truncation act on the mixture and need both projections.

\paragraph{Gradient paths and baseline containment.} Each table receives its own output-projection gradient plus half of the gradient through $\mM=(\mA+\mB)/2$, which covers both the decoder input and the token gate, since the gate input is not detached. With $\mA=\mB=\mE$, both output distributions equal the tied baseline's and the mixture no longer depends on the gate, so copying a tied checkpoint's table into both tables and keeping its decoder preserves its predictions. The reported experiments do not use this construction; they initialize the two tables independently (Appendix~\ref{app:setup}).

\paragraph{Token routing and the MoS comparison.} The gate learns one mixture weight per previous-token type, using token identity as in hash-based routing~\citep{roller2021hash} but with learned soft weights instead of fixed assignments. MoS~\citep{yang2018breaking} uses $K$ transformed hidden states under one tied table with a context-dependent prior; \gate{} uses one hidden state, two tables and a token-dependent prior. At $K=2$ the MoS transformation matrices add $2d^2$ parameters, against $Vd$ for the second table, plus the gate parameters in each design. The gradient paths also differ: MoS sends every component's gradient into the one input table, whereas each table in \gate{} gets its own output-projection gradient.

\paragraph{Training overhead.} The second $V\times d$ table adds 38.6M parameters at 142M, 51.5M at 330M and 103M at 1B. Training also materializes two fp32 log-probability vectors per position. In the 1B Dense timing run on 8$\times$H100, throughput is 443K tokens/s for \gate{} versus 502K for Standard, a 12\% reduction.

\paragraph{Alternative output layers and their implementation.} Section~\ref{sec:output-comparison} adapts the published output formulas to the 142M Dense Transformer. MoS uses two transformed states, $\tanh(W_kh)$, a shared tied vocabulary table and a hidden-state-dependent mixture prior. Sigsoftmax uses $p_v\propto\exp(z_v)\sigma(z_v)$ without adding parameters. MoSS applies this normalization within each of two MoS components. The untied head uses independent input and output tables to control for the additional vocabulary parameters. All output projections are bias-free.

Each method is trained from scratch with seed 0 and the same initial body and input-table weights, data order, optimizer schedule and validation windows as the factorial experiment in Section~\ref{sec:factorial}. We reuse its Standard and \gate{} runs. Training uses two H100 GPUs, 16 sequences per GPU per microbatch and eight accumulation steps, for 21,740 updates and 5.699B tokens (40 TPP). We did not tune a separate recipe per method, and we did not reproduce the original papers' recurrent-model experiments. All six final checkpoints are evaluated on the full perplexity test sets and the eight-task suite of Tables~\ref{tab:main} and~\ref{tab:downstream}, with the same prompts and scoring rules.

\paragraph{Cost measurement for alternative heads.} Table~\ref{tab:output-comparison} reports median aggregate training throughput across the two GPUs, excluding validation and checkpoint writes. Memory is measured as peak PyTorch-allocated memory on rank 0 in GB. All measurements use the hardware and batch configuration above. Training tokens are matched across methods, while parameter counts and FLOPs differ.

\paragraph{Factorial ablation and its protocol.} Section~\ref{sec:factorial} uses a $2\times2$ grid of 142M Dense models with about 104M non-embedding parameters. Let $q_A(h)=\operatorname{softmax}(\mA h)$ and $q_B(h)=\operatorname{softmax}(\mB h)$. The input is either $\mA[x_t]$ or $\mM[x_t]=(\mA[x_t]+\mB[x_t])/2$, and the output is either $q_A$ or $g_tq_A+(1-g_t)q_B$. Both mixture-output cells use $g_t=\sigma(\vw^\top\operatorname{RMSNorm}(x_t^{(0)}))$, where $x_t^{(0)}$ is the input vector for that cell. Depending on the cell, $\mB$ enters the input, the output, both or neither. The input-only cell combines fixed averaging with the $\mA$ head. This differs from the separate input-gate control, which reads $g_v\mA_v+(1-g_v)\mB_v$ and uses $\mM$ at the head.

All four cells are trained from scratch with seed 0, the same data order, optimizer schedule and 40-TPP token budget, and the same 1,024 validation windows. Each uses two H100 GPUs, a global batch of 262,144 tokens and 21,740 updates. We initialize the complete two-table model before removing unused parameters, preserving identical initial body and $\mA$ weights across cells. Comparisons use this grid's Standard baseline rather than the primary baseline in Table~\ref{tab:main}.

\paragraph{Depth-control budget.} At 330M, the second vocabulary table contains $Vd=51.5$M parameters, while each Transformer layer contains approximately 11.6M. Adding four layers, from 24 to 28, gives a similar but not identical parameter increase and roughly matches the added training FLOPs. Section~\ref{sec:depth-control} uses the separate depth-control group in Table~\ref{tab:controls}. No corresponding 1B experiment has been run.

\paragraph{Downstream protocol.} We evaluate on HellaSwag~\citep{zellers2019hellaswag}, ARC-Easy/Challenge~\citep{clark2018think}, PIQA~\citep{bisk2020piqa}, BoolQ~\citep{clark2019boolq}, SciQ~\citep{welbl2017crowdsourcing}, LAMBADA~\citep{paperno2016lambada}, and MMLU~\citep{hendrycks2021measuring}. Our evaluator follows lm-evaluation-harness~\citep{gao2024harness} with a few differences. HellaSwag and ARC scores labeled acc\_norm use the UTF-8 byte length of the continuation, including its prepended space, rather than the harness's character-length denominator. HellaSwag preprocessing normalizes whitespace but omits some harness transformations, including bracket removal and capitalization of the second context fragment. MMLU prompts contain five development examples per subject before tokenization. Sequences longer than 1,024 tokens are then left-truncated, which can remove the header and early examples. The retained context is therefore not always five-shot. Our scores are not directly comparable with published harness numbers.

\paragraph{Layouts and hardware.} The 142M Standard runs used one H100 with micro-batch 32 and accumulation 8. The 142M \gate{} runs used eight GPUs with micro-batch 16 and accumulation 2, except for the MoE \gate{} run, which used four B200 GPUs with micro-batch 16 and accumulation 4. All used 262,144 tokens per step in the same window order. The 330M runs generally used four H100 GPUs with micro-batch 16 and accumulation 4. The exceptions were Dense \gate{}, which used eight GPUs and accumulation 2, and the MoE pair, whose two arms used four B200 GPUs with the same layout.

The 1B runs used eight GPUs with micro-batch 8 and accumulation 8. After a node failure at step 28,000, the 1B Standard run moved from H100 to B200 and resumed from its checkpoint. The checkpoint's validation loss reproduced to $10^{-4}$ on the new hardware. MoE runs under PyTorch 2.7.1 showed loss spikes for every twin variant and are omitted. The MoE runs in Table~\ref{tab:main} used a newer PyTorch version and trained smoothly.

\section{Setup and hyperparameters}
\label{app:setup}

\paragraph{Decoder and tokenizer details.} The decoders use pre-norm RMSNorm~\citep{zhang2019root}, SwiGLU~\citep{shazeer2020glu}, grouped-query attention~\citep{ainslie2023gqa} with QK-norm~\citep{henry2020querykey}, and RoPE~\citep{su2024roformer}. Training uses no dropout. GPT-2 BPE has 50,257 tokenizer entries, and one end-of-text token is appended to each document. The embedding and output tables have $V{=}50{,}304$ rows.

\begin{table}[!htbp]
\centering
\small
\caption{\textbf{Hyperparameters.} All sizes share the recipe except where stated.}
\label{tab:hparams}
\twdFit{\begin{tabular}{@{}p{2.3cm}p{4.6cm}p{2.0cm}p{4.6cm}@{}}
\toprule
\multicolumn{2}{@{}l}{\textit{Model}} & \multicolumn{2}{l@{}}{\textit{Training}} \\
\cmidrule(r){1-2}\cmidrule(l){3-4}
Layers / width & 16 / 768 (142M); 24 / 1,024 (330M); 20 / 2,048 (1B) & Optimizer & AdamW, $\beta{=}(0.9,0.95)$, $\epsilon{=}10^{-8}$ \\
Heads (query / KV) & 12 / 6 and 16 / 8, head dim 64; 1B: 16 / 8, head dim 128 & Learning rate & $10^{-3}$ peak (1B: $6\times10^{-4}$), 1,000 warmup steps (1B: 2,000) \\
MLP & SwiGLU, $64\lceil\lfloor\tfrac83 d\rfloor/64\rceil$ hidden (1B: 5,504) & Schedule & WSD: constant, then linear over the last 20\% \\
Vocabulary & 50,304 (GPT-2 BPE) & Weight decay & 0.1 on every matrix incl.\ the tables \\
Context & 1,024 tokens, RoPE $\theta{=}10^6$ & Clipping & 1.0 (global norm) \\
Normalization & pre-norm RMSNorm, QK-norm & Batch & 262,144 tokens per step (1B: 524,288) \\
Tying & input and output tables tied (both tables for \gate{}) & Budget / steps & 40 TPP; 21,740 steps (142M), 50,349 (330M), 78,800 (1B) \\
Initialization & $\mathcal N(0,0.02^2)$; $W_O$, $W_2$: $0.02/\sqrt{2L}$ & Precision & bf16 autocast, fp32 master weights, \texttt{torch.compile} \\
GDN hybrid & 3 Gated DeltaNet : 1 attention, head dim 128, conv 4 & Evaluation & every 500 steps (1B: 1,000); primary 142M/330M: 1,024 fixed windows \\
MoE & 8 experts of width $d_{\mathrm{ff}}/2$, top-2, capacity 1.25, aux 0.01 & Data & 7.18B-token pool (142M, 330M), 23.9B-token disjoint pool (1B) \\
AttnRes & Full Attention Residuals, block size 0 & & \\
\bottomrule
\end{tabular}}
\end{table}

\paragraph{Table initialization.} The baseline table $\mE$ and each of $\mA,\mB$ use entrywise $\mathcal N(0,0.02^2)$ initialization, with $\mA$ and $\mB$ drawn independently. Thus $\operatorname{Var}(M_{vi})=(\operatorname{Var}(A_{vi})+\operatorname{Var}(B_{vi}))/4=0.02^2/2$, so the averaged input table has standard deviation $0.02/\sqrt{2}$. The reported experiments do not use the baseline-preserving construction $\mA=\mB=\mE$.

Table~\ref{tab:hparams} lists the model and training configurations. At 142M scale, the non-embedding parameter counts are 103.8M for Dense, 118.2M for GDN hybrid, 330.4M total and 103.9M active for MoE, and 103.9M for AttnRes. The Dense counts at 330M and 1B are 278.5M and 928M.

All validation windows are drawn from a common 10M-token held-out pool. The separate control and additional evaluations sample 50 batches with a fixed seed, using the training micro-batch size. This gives 1,600 windows for 142M Standard, 800 for the 142M twin-table and control runs, 800 for the reported 330M pairs and 400 for the 1B pair. DDP ranks evaluate the same windows rather than adding independent samples. This variation in window count affects only the separate control and additional evaluations; the primary 142M and 330M comparisons use the common-window protocol below.

\paragraph{Training budget and primary evaluation protocol.} The 1B runs use 78,800 steps at 524,288 tokens per step, totaling 41,313,894,400 tokens, or approximately 41.31B. This exceeds the 23.9B-token pool, so the data is repeated. The step counts and common-window evaluation protocol below apply to the primary 142M and 330M runs.

For these runs, the token budget is 40 times the Dense Standard parameter count (142,469,888 and 329,961,472), rounded up to whole steps. A global batch of 256 sequences of length 1,024 gives 262,144 tokens per step. Training therefore lasts 21,740 steps and 5,699,010,560 tokens at 142M, and 50,349 steps and 13,198,688,256 tokens at 330M. All arms at each size share this budget, including the larger twin-table models and MoE models with additional expert parameters. We traverse the 7,178,747,873-token pool in non-overlapping windows shuffled with seed 0. The same permutation repeats when the pool is exhausted, giving approximately 1.84 passes at 330M. AdamW uses $\beta=(0.9,0.95)$, weight decay 0.1, peak learning rate $10^{-3}$ and 1,000 warmup steps. The learning rate decreases linearly to $10^{-4}$ over the last 20\% of training. We use gradient clipping at 1, bf16 and compilation.

Every arm evaluates the same 1,024 held-out windows sampled with seed 1234, in batches of eight, every 500 steps and at the final step. GPU layouts and software versions vary across runs, and we have no multi-seed uncertainty estimate. Table~\ref{tab:main} reports final-checkpoint losses rather than intermediate best-checkpoint losses.

We evaluate LAMBADA and WikiText-2 perplexity on all sixteen final checkpoints. LAMBADA uses all 5,153 test examples, and WikiText-2 uses all 62 test documents. Token, word and byte counts are identical across arms. Scoring uses the GPT-2 tokenizer, bf16, rolling 1,024-token windows and the served model, with its averaged input table and token-gated output mixture. MoE inference uses capacity 4.0 to avoid dropping tokens.

The A-only and B-only evaluations use the final twin checkpoints, the same 1,024 validation windows and batch size eight. They retain the training-time rank-strided batch grouping, which affects MoE capacity. Both input and output use the selected table without a gate. MoE capacity remains 1.25, as in training-time validation, rather than the 4.0 used for perplexity evaluation. As a consistency check, the unablated mixture reproduces the logged final loss for all eight checkpoints.

Table~\ref{tab:downstream} evaluates all sixteen final checkpoints on the complete eight-task suite. The sample counts are 10,042 for HellaSwag, 2,376 for ARC-Easy, 1,172 for ARC-Challenge, 1,838 for PIQA, 3,270 for BoolQ, 1,000 for SciQ, 5,153 for LAMBADA and 14,042 for MMLU. Downstream scoring uses the same tokenizer, precision and served model, with MoE capacity 4.0. LAMBADA accuracy reuses the greedy predictions from the perplexity evaluation of the same checkpoint. MMLU uses up to five shots with the truncation described in Appendix~\ref{app:code}. The reported average weights all eight tasks equally, and task samples and order are shared across arms.

\paragraph{Run-to-run variation.} In the separate control experiments, two 142M Standard seeds finish 0.008 apart (2.8777 and 2.8693). At 330M, two Standard runs differing only in GPU layout finish 0.001 apart. These are the only repeats we have; they are not uncertainty estimates for the primary single-seed results. The tables mark with $^\ddagger$ the runs that use a different number of GPUs from their Standard counterpart, with the same global batch and window order.

\section{Variants by their name in the code}
\label{app:variants}

\begin{table}[!htbp]
\centering
\footnotesize
\caption{\textbf{\gate{} and the controls} by their \texttt{embed\_mode} in the code. $\mM=\tfrac12(\mA+\mB)$.}
\label{tab:variants}
\twdFit{\begin{tabular}{@{}p{0.2\linewidth}p{0.25\linewidth}p{0.4\linewidth}p{0.09\linewidth}@{}}
\toprule
Model & Code & Input; head & Serving \tabularnewline
\midrule
Standard & \texttt{single} & $\mE$; $\mE$ (tied) & 1 table \tabularnewline
Standard, 28 layers & \texttt{single}, \texttt{n\_layer=28} & $\mE$; $\mE$ & 1 table \tabularnewline
Untied head & \texttt{single}, \texttt{tie\_\allowbreak embeddings=false} & $\mE$; separate $\mW$ & 2 tables \tabularnewline
MoS, $K{=}2$ & \texttt{mos2} & $\mE$; $\sum_k\pi_k(\vh)\operatorname{softmax}(\mE\tanh(\mW_k\vh))$ & 1 table \tabularnewline
Input gate only & \texttt{gin\_only} & $g_v\mA_v+(1{-}g_v)\mB_v$; $\mM$ & 2 tables \tabularnewline
\midrule
\gate{} & \texttt{router\_mos\_tok} & $\mM$; $g\operatorname{softmax}(\mA\vh)+(1{-}g)\operatorname{softmax}(\mB\vh)$, $g=\sigma(\vw^\top\operatorname{rmsnorm}(\mM_{x_t}))$ & 2 tables \tabularnewline
\bottomrule
\end{tabular}}
\end{table}

\section{Limitations}
\label{app:limitations}

Each arm has one seed, so we cannot estimate seed-to-seed variance, even though the primary loss comparisons share validation windows. The separate 142M controls and additional evaluations use unequal numbers of validation windows, and hardware, GPU layout and software versions vary across runs. None of the downstream averages has a confidence interval or a replicate seed.

Equal token budgets do not match parameters or FLOPs, since \gate{} adds a table and a projection; the roughly resource-matched depth control exists at 330M only. Width and depth change together between 142M and 330M, and the larger budget repeats the shuffled data, so the runs do not separate width, training length and data repetition. At 1B, MoE has worse loss and perplexity but better downstream accuracy. The 1B runs use separate validation windows, and we have no explanation for the disagreement.

The baseline and twin models also start from input tables of different scale, and no control isolates that difference. The one-table ablations change input and output together, so their loss penalties show neither output-only specialization nor the source of the gain over Standard. We have not tested larger vocabularies or longer contexts, and we have not looked at which token types the gate assigns to which table.

\section{Additional evaluations}
\label{app:legacyresults}
Table~\ref{tab:legacy20} records an earlier round of evaluations. These used different evaluation settings (Appendix~\ref{app:setup}), including different numbers of validation windows across the 142M arms, and are superseded by Table~\ref{tab:main}.
\begin{table}[!htbp]
\centering
\caption{\textbf{Additional evaluations} using earlier settings. These results are not comparable with Table~\ref{tab:main}, although the column definitions are the same.}
\label{tab:legacy20}
\begingroup\footnotesize\setlength{\tabcolsep}{2.4pt}
\begin{tabular}{@{}llccccccccc@{}}
\toprule
 & & \multicolumn{3}{c}{Val loss} & \multicolumn{2}{c}{LAMBADA PPL} & \multicolumn{2}{c}{WikiText PPL} & \multicolumn{2}{c}{One table} \\
\cmidrule(lr){3-5}\cmidrule(lr){6-7}\cmidrule(lr){8-9}\cmidrule(lr){10-11}
Size & Arch. & Std. & \gate{} & $\Delta$ & Std. & \gate{} & Std. & \gate{} & $\mA$ & $\mB$ \\
\midrule
142M & Dense & 2.8777 & \textbf{2.8501}$^\ddagger$ & \textbf{$-$0.028} & 87.3 & \textbf{71.8} & 72.1 & \textbf{68.2} & 3.27 & 3.17 \\
 & GDN hybrid & 2.8208 & \textbf{2.8101}$^\ddagger$ & \textbf{$-$0.011} & 68.2 & \textbf{65.0} & 64.6 & \textbf{64.1} & 3.20 & 3.12 \\
 & MoE & 2.7890 & \textbf{2.7775}$^\ddagger$ & \textbf{$-$0.011} & \textbf{62.0} & 63.0 & 62.3 & \textbf{60.2} & 3.00 & 3.66 \\
 & AttnRes & 2.8493 & \textbf{2.8208}$^\ddagger$ & \textbf{$-$0.029} & 78.5 & \textbf{65.2} & 65.6 & \textbf{62.5} & 3.25 & 3.26 \\
\addlinespace
330M & Dense & 2.6426 & \textbf{2.6329}$^\ddagger$ & \textbf{$-$0.010} & 35.0 & \textbf{34.4} & 47.4 & \textbf{46.9} & 2.92 & 2.88 \\
 & GDN hybrid & 2.5989 & \textbf{2.5930} & \textbf{$-$0.006} & 31.9 & \textbf{29.0} & 44.3 & \textbf{43.9} & 2.78 & 2.92 \\
 & MoE & 2.5769 & \textbf{2.5625} & \textbf{$-$0.014} & 31.7 & \textbf{27.6} & 43.2 & \textbf{41.6} & 2.83 & 3.69 \\
 & AttnRes & 2.6150 & \textbf{2.6057} & \textbf{$-$0.009} & 33.1 & \textbf{29.6} & 44.0 & \textbf{42.9} & 2.88 & 2.84 \\
\addlinespace
1B & Dense & 2.3875 & \textbf{2.3766} & \textbf{$-$0.011} & 16.1 & \textbf{15.8} & 30.2 & \textbf{29.7} & 3.10 & 2.53 \\
\bottomrule\end{tabular}\endgroup

\end{table}

\end{document}